\documentclass[onefignum,onetabnum]{article}
\usepackage[switch, running]{lineno}

\usepackage[top=1in, bottom=1in, left=1in, right=1in]{geometry}
\usepackage[title]{appendix}
\usepackage{authblk}
\usepackage{amsmath,amsfonts,amscd,amssymb}
\usepackage{graphicx}
\usepackage{epstopdf}
\usepackage{overpic}
\usepackage{cancel}
\usepackage{rotating}
\usepackage{url}
\usepackage{caption}
\usepackage{color}
\usepackage{rotating}
\usepackage{multirow}
\usepackage{wrapfig}
\usepackage{mathtools}
\usepackage{subeqnarray}
\usepackage{setspace}
\usepackage{standalone}
\usepackage{subfig}
\usepackage[normalem]{ulem}
\usepackage{tikz}
\usepackage{palatino} % needs 10
\usepackage[numbers,sort&compress]{natbib}
\usepackage{hyperref}

\usepackage[bottom,flushmargin,hang,multiple]{footmisc}
\usepackage{lipsum}

\definecolor{header1}{cmyk}{0,0,0,1}

\DeclareMathOperator*{\argmin}{arg\rm{}min}

\title{Discrepancy Modeling Framework: Learning missing physics, modeling systematic residuals, and disambiguating between deterministic and random effects}

\author[1]{Megan R. Ebers}
\author[1]{Katherine M. Steele}
\author[2]{J. Nathan Kutz}
\affil[1]{Department of Mechanical Engineering, University of Washington}
\affil[2]{Department of Applied Mathematics, University of Washington}

\begin{document}

\maketitle

%\linenumbers 
 
\begin{abstract}
Physics-based and first-principles models pervade the engineering and physical sciences, allowing for the ability to model the dynamics of complex systems with a prescribed accuracy.  
The approximations used in deriving governing equations often result in discrepancies between the model and sensor-based measurements of the system, revealing the approximate nature of the equations and/or the signal-to-noise ratio of the sensor itself. 
In modern dynamical systems, such discrepancies between model and measurement can lead to poor quantification, often undermining the ability to produce accurate and precise control algorithms.  
\textcolor{black}{We introduce a discrepancy modeling framework to identify the missing physics and resolve the model-measurement mismatch with two distinct approaches: (i) by learning a model for the evolution of systematic state-space residual, and (ii) by discovering a model for the deterministic dynamical error.}
Regardless of approach, a common suite of data-driven model discovery methods can be used. Specifically, we use four fundamentally different methods to demonstrate the mathematical implementations of discrepancy modeling: (i) the {\em sparse identification of nonlinear dynamics} (SINDy), (ii) {\em dynamic mode decomposition} (DMD), (iii) {\em Gaussian process regression} (GPR), and (iv) {\em neural networks} (NN).
\textcolor{black}{The choice of method depends on one's intent (\textit{e.g.}, mechanistic interpretability) for discrepancy modeling, sensor measurement characteristics (\textit{e.g.}, quantity, quality, resolution), and constraints imposed by practical applications (\textit{e.g.}, state- or dynamical-space operability).}
We demonstrate the utility and suitability for both discrepancy modeling approaches using the suite of data-driven modeling methods on three continuous dynamical systems under varying signal-to-noise ratios.
\textcolor{black}{Finally, we emphasize structural shortcomings of each discrepancy modeling approach depending on error type. In summary, if the true dynamics are unknown (\textit{i.e.}, an imperfect model), one should learn a discrepancy model of the missing physics in the dynamical space.
Yet, if the true dynamics are known yet model-measurement mismatch still exists, one should learn a discrepancy model in the state space.}
\end{abstract}

\section{Introduction}

The traditional modeling of physics and engineering systems relies on the development of governing equations that characterize the underlying nonlinear, dynamical processes.  Such governing equations are typically derived through asymptotic reductions, enforcing physical constraints or conservation laws, and/or positing empirical relations between variables~\cite{lin1988mathematics}. Simulation of the governing equations then allows for prediction, control, and characterization of the complex system.  However, it is well known that governing equations are often idealized and only approximate, either achieved through dominant balance physics arguments or neglecting higher-order effects~\cite{bender2013advanced,kutz2020advanced,callaham2021learning,bakarji2022dimensionally}. In many emerging fields, the idealized models currently used are simply inadequate for modeling applications where precision is necessary, such as in robotics, biomechanics, precision manufacturing, and automated systems. The time evolution of complex nonlinear systems is often highly sensitive to small errors in system dynamics. This limits the utility of simulation, such as for control or inference. \textcolor{black}{Generally, there are two kinds of errors that occur in modeling physical systems:  missing physics and measurement error (which can be systematic or random).} In practice, both errors exist and are difficult to disambiguate. \textcolor{black}{Our framework provides approaches which help disambiguate between the dominant error forms to estimate the missing physics, either by learning the deterministic dynamical error or characterizing the state-space residual between model and measurement.}

With the advancement of modern sensor technologies, there is opportunity to improve the characterization of system dynamics through modern data-driven methods to refine and augment the known, governing first-principles. Indeed, a better understanding of the underlying physical processes can be achieved by inspecting the error between first-principles theory and sensor measurements of dynamical systems. The error may contain deterministic effects, or {\em discrepancies}, and models of the discrepancy can be learned using data-driven model discovery. 
A number of machine learning techniques have been developed to learn model error or discrepancies using hybrid data assimilation techniques, \textcolor{black}{in which additive correction models of the missing physics}~\cite{saveriano2017data,harlim2021machine,farchi2021using,shi2019neural} are learned for a diversity of applications.
\textcolor{black}{For example, Levine and Stuart~\cite{levine2021framework} recently proposed leveraging machine learning methods, specifically recurrent neural networks, for modeling hybrid problems in non-Markovian settings.} By augmenting known first-principles with a discrepancy model estimating the missing physics, an improved dynamical model can be learned. 
\textcolor{black}{In certain practical applications, such as in controls engineering~\cite{aastrom2021feedback, nise2020control, golnaraghi2017automatic, ogata1978system, ogata2010modern}, the dynamical model may be unavailable or infeasible to interface with, thereby necessitating learning a discrepancy model of the state-space residual to correct system approximations.}
Although we have always improved our models through systematic approaches, we build on recent work~\cite{kaheman2019learning,levine2021framework} and outline here two principled, data-driven approaches by which discrepancy modeling is automated. Moreover, we examine the relationship between deterministic and random effects within the model-measurement mismatch; experimental noise in sensor measurements dictate limitations in learning a discrepancy model.

Discrepancy modeling has deep historical context, especially since early models of any system are typically coarse approximations of the physics. Indeed, throughout the 1960s, before the advancement of scientific computing, asymptotic and perturbation methods~\cite{bender2013advanced,kutz2020advanced} that systematically introduced discrepancies were leading methods in the development of fluid dynamics. From the Prandtl number to the Reynolds number, various approximations led to different dominant balance physics approximations~\cite{batchelor2000introduction,callaham2021learning,bakarji2022dimensionally}. Thus by construction, such models allowed for improved understanding at the expense of detailed models. Computation has allowed us to move beyond such reductions, yet model-measurement mismatch continues to prevent accurate interpretation of system structure and/or prediction of time evolution. 
{\color{black}
In modern applications, discrepancy modeling can play a foundational role in improved modeling across any data-driven application.  As an example, consider emerging digital twin technologies which require a computational model of a real system.  Such models are typically idealized from first principle physics, which fail to accurately match reality.  This is especially problematic where precision is required in the digital twin.  Thus almost any practical systems of interest can benefit from a discrepancy modeling improvement and evaluation.  Kennedy and O'Hagan~\cite{kennedy2001bayesian} provide a thorough review of uncertainty quantification techniques from a statistical perspective.  The current methods advocated, which are focused on dynamical systems, give additional techniques which can be used for improved modeling purposes.}
Indeed, in almost every application area, mismatch exists between experiment and theory which is not due to noise. For instance, improvements in tracking planetary motion in the late 1800s and early 1900s allowed for the characterization of a discrepancy between Newton's gravitation laws and the observed physics, eventually leading to the development of general relativity by Einstein~\cite{wald2010general}. More recently, identifying missing deterministic effects (provided they are not obscured by noise) has allowed for the discovery of missing physics that are challenging to model with first principles, such as fluid drag forces of falling~\cite{de2020discovery} or orbiting~\cite{manzi2020discovering} objects or with bearing chatter during double pendulum control~\cite{kaheman2019learning}. Our goal is to leverage improved sensor observations; we automate the process of building better models by identifying a discrepancy model.  
{\color{black} Note that this is a significantly different task than characterizing the sensitivity of the system to initial condition which is a hallmark feature of chaotic systems.}

The ideas in our discrepancy modeling framework are well established concepts in \textcolor{black}{statistics, optimization, and controls}~\cite{bevington2003data, taylor1997introduction, cook1982residuals, cox1968general, box1970distribution, welch1995introduction}. \textcolor{black}{In statistical regression analysis}, measures of deviation are referred to as either {\em error} or {\em residual}. An {\em error} is the difference between the observed values and the true values. A {\em residual} is the difference between the observed values and the estimated values. In practice, observed values are often used as a proxy for the true values; therefore, residuals may contain both random and deterministic signals. Regression analysis seeks to evaluate how well a statistical model fits a data set; if the residual contains a bias, it suggests the model can be improved by capturing deterministic values within the residual. We posit that discrepancy modeling is to dynamical systems modeling what regression analysis is to statistical modeling. 
\textcolor{black}{In controls engineering}, traditional approaches to discrepancy modeling in a state-space representation include Kalman filtering for data assimilation; however, \textcolor{black}{the dominant assumption is that} the mismatch between model and measurement are given by normally-distributed variables, \textit{i.e.}, random processes~\cite{welch1995introduction, law2015data}. \textcolor{black}{Therefore, the state-space model is not updated through self-improvement to account for non-random processes. Whether modeling a dynamic system in a dynamical or state space, the error may reveal deterministic structure affecting the system's observed time evolution}. As already noted, Levine and Stuart~\cite{levine2021framework} recently have advanced the state-of-the-art by leveraging machine learning methods (neural networks) for learning dynamical error. More broadly, they provide a rigorous analysis for learning such models from data \textcolor{black}{in a dynamical representation, \textit{e.g.}, differential equations}. In this work, we build on this theme by considering a broader class of models, including those constructed by {\em sparse identification of dynamical systems} (SINDy), {\em dynamic mode decomposition} (DMD), and {\em Gaussian process regression} (GPR). \textcolor{black}{We also advocate for learning a discrepancy model of the systematic residuals in the state space}. Again, we demonstrate SINDy, DMD, and GPR along with neural networks (NN) to learn models of the systematic residual directly and correct model-measurement mismatch in the state space. 
This hybrid (mechanism+data) framework brings together domain knowledge from first principles and data-driven model discovery to provide a more comprehensive modeling space~\cite{miller2021breiman, smith2013uncertainty}. 

\begin{table}[t]
\centering
 \begin{tabular}{||l l||} 
 \hline
 Variable & Description \\ [0.5ex] 
 \hline\hline
 ${\bf x} (t) \in \mathbb{R}^n$ & approximate state space  \\ 
 ${\tilde{\bf x}} (t) \in \mathbb{R}^n$ & augmented state space  \\
 ${\bf x}_0 (t) \in \mathbb{R}^n$ & true state space  \\
 ${\bf y}_k = {\bf y}(k \Delta t)  \in \mathbb{R}^n$ & measurements  \\
 & \\
 $f(\cdot)$ & approximate dynamics  \\
 $\color{black}{g}(\cdot)$ & \textcolor{black}{missing physics} \\
  $F(\cdot)$ & true dynamics  \\
 $\tilde{F}(\cdot)$ & augmented dynamics  \\ 
  & \\
 ${\textcolor{black}{{\bf x}_D}(t_k)} = {\bf y}_k - {\bf x}(t_k) $ & state-space residual  \\
 ${\textcolor{black}{{\bf \dot{x}}_D}(t_k)}= {\bf \dot{y}}_k - f({\bf y}_k) $ & deterministic dynamical error \\
 & \\
 $\textcolor{black}{{\delta}(\cdot)}$ & \textcolor{black}{discrepancy model estimating the state-space residual} \\
 $\textcolor{black}{{\delta}_f(\cdot)}$ & \textcolor{black}{discrepancy model estimating the deterministic dynamical error} \\
 [1ex] 
 \hline
 \end{tabular}
 \caption{Variable definitions. \textcolor{black}{The goal of discrepancy modeling is to improve the Platonic or idealized model via augmentation of a learned discrepancy model, such that the model solution and measurement data converge.}  \textcolor{black}{There are two approaches to build a discrepancy model and estimate missing physics in a dynamical system: learn the deterministic dynamical error or model the systematic state-space residual.} \\
 \label{tbl:abrv}}
\end{table}

From a modeling perspective (see definitions in Table~\ref{tbl:abrv}), it is assumed that approximate dynamics of the system are known, \textit{i.e.,} the Platonic model:
\begin{equation}
    {\dot{\bf x}}(t) = f({\bf x}(t))
  \label{eq:plato}
\end{equation}
\textcolor{black}{where ${\bf x} (t) \in \mathbb{R}^n$ and} the approximate governing dynamics $f(\cdot)$ are known and derived from first principles, asymptotic reductions, enforcing physical constraints or conservation laws, or positing empirical relations between variables. However, in truth, the true dynamics $F(\cdot)$, {\em which we do not have access to}, are given by
\begin{equation}
    {\dot{\bf x}_0}(t) = F({\bf x}_0(t))= f({\bf x}_0(t)) + g({\bf x}_0(t)) %{\color{black}{+ {\cal N}_1(\mu_1,\sigma_1) }}
    \label{eq:truth}
\end{equation}
\textcolor{black}{where ${\bf x}_0(t) \in \mathbb{R}^n$ and} $g(\cdot)$ is the missing (deterministic) physics that remains unmodeled due to some suitable approximation and/or lack of physics knowledge. Note that the missing physics comprising $g(\cdot)$ may be intentionally omitted (\textit{e.g.}, through model reduction) or unintentionally omitted (\textit{e.g.}, due to lack of first-principles knowledge of the system structure). \textcolor{black}{While not discussed in this manuscript, there may exist a noise process that drives stochastic variability in the model. If the stochastic process also remains unmodeled, the residual becomes the combination of the missing deterministic and stochastic physics. Therefore, those two terms would need to be disambiguated in order to learn a discrepancy model of the deterministic effect.}

\textcolor{black}{A time series of the system's full state is observed at discrete time points so that}
\begin{equation}
  {\bf y}_k = {\bf x}_0(t_k)
  + \color{black}{{\cal N}(\mu,\sigma)}
  \label{eq:measurements}
\end{equation}
\textcolor{black}{where ${\bf y}_k \in \mathbb{R}^n$ and ${\cal N}(\mu,\sigma)$ is a noise process describing observation (or sensor) noise. The processes driving the observed dynamical systems are often continuous, and measurements can be collected at regular time intervals $\Delta t$ via the observation process in Eq.~\ref{eq:measurements}, such that ${\bf y}_k = {\bf y}(k \Delta t)$.}

{\color{black}In this manuscript, fairly basic assumptions are made concerning the noise. We will consider Gaussian additive noise in the sensor that is parameterized by a mean and variance. For sufficiently large noise, or small signal-to-noise ratio, there are few methods capable of producing viable discrepancy models.  Of course, smoothing of the data can be attempted, but for large noise such signal filtering can also compromise any ability to identify a reasonable discrepancy model.  Thus we consider noise levels that are low or moderate in order to apply the model discovery methods advocated.  For large noise, Gaussian process regression is perhaps the best technique available. This is an important consideration, as this formulation relies on the ability to compute a good approximation of derivatives from the observed measurements.}

\textcolor{black}{The goal of discrepancy modeling is to resolve model-measurement mismatch in dynamical systems. This can be done in two distinct ways; a similar distinction has been discussed in \cite{farchi2021using}.} First, one can generate an improved dynamical model:
\begin{equation}
        {\dot{\tilde{\bf x}}}(t) = \tilde{F} (\tilde{\bf x}(t)) = f(\tilde{\bf x}(t)) + \color{black}{{\delta}_f}(\tilde{\bf x}(t)),
        \label{eq:learning}
\end{equation}
by learning a discrepancy model of the deterministic dynamical error (\textit{e.g.}, missing physics), $\delta_f (\cdot)$, and appending it to the known imperfect/approximate dynamics, such that $\color{black}{\|} {\bf y}_k - \tilde{\bf x}(t_k) \color{black}{\|} < \color{black}{\|} {\bf y}_k - {\bf x}(t_k) \color{black}{\|}$, \textit{i.e.}, the augmented model's system estimations are less erroneous than the approximate model's. \textcolor{black}{In this case, because our goal is to recover the missing physics to resolve model-measurement mismatch, priority should be given to learning a discrepancy model in the dynamical space.} 

\begin{figure}[t]
    \raggedleft
    \includegraphics[width=\textwidth]{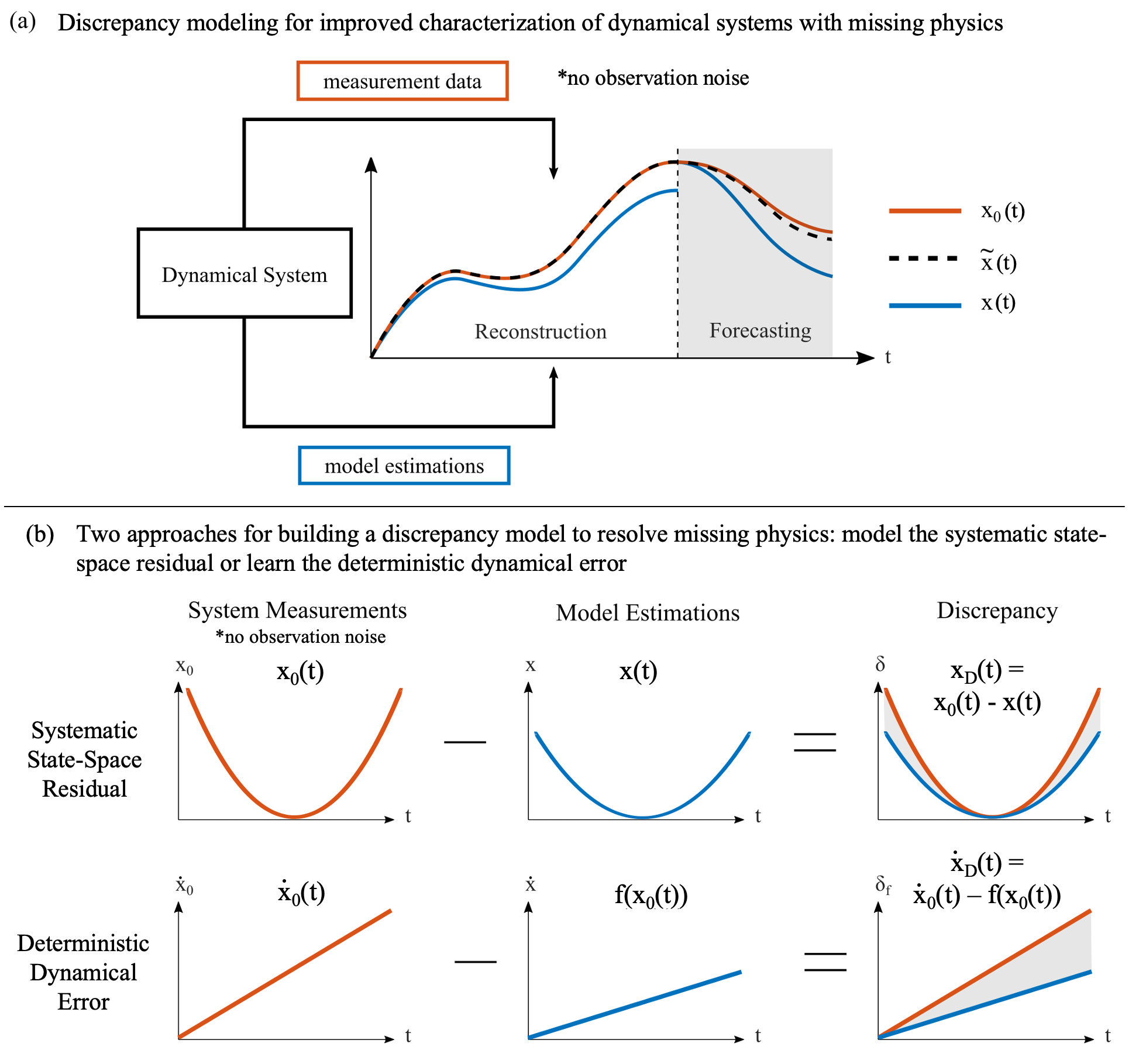}
    \caption{\textbf{Top panel}: An approximate dynamical model $f(\cdot)$ provides estimates of system behavior {\color{black}used for both reconstruction and forecasting (shaded region)}, ${\bf x}(t)$. However, true behavior ${\bf x_0}(t)$ (without observation noise) deviates from these estimates. The goal of discrepancy modeling is to learn a discrepancy model that recovers the missing physics and augments the approximate dynamics to improve system characterization, $\tilde{\bf x}(t)$. \textbf{Bottom panel}: \textcolor{black}{There are two approaches for building a discrepancy model to estimate missing physics: (i) modeling systematic state-space residual between the approximate state space, ${\bf x}(t)$, and true state space, ${\bf x}_0(t)$, and (ii) learning the deterministic dynamical error between the true dynamics, $\dot{{\bf x}}_0(t) = f({\bf x}_0(t)) + g({\bf x}_0(t))$, and the approximate dynamics, $\dot{\bf x}(t) = f({\bf x}(t))$. In real-world systems, the true system behavior is noisily observed, ${\bf y}_k = {\bf x}_0(t_k) + \mathcal{N}(\mu, \sigma)$, model-measurement mismatch contains both deterministic and random effects; measurements ${\bf y}_k = {\bf y}(k \Delta t)$ denote a continuous dynamical system's full state noisily observed at discrete time points.}}
    \label{fig:MainFigure}
\end{figure}

\textcolor{black}{In the second case, if the true dynamics are known yet a model-measurement mismatch still exists, one can decrease the mismatch by learning the systematic residual (\textit{e.g.}, observation error)} \textcolor{black}{\cite{kennedy2001bayesian}}:
\begin{equation}
       {\bf \tilde{x}}(t) = {\bf x}(t) + \color{black}{{\delta}}({\bf x}(t)) .
       \label{eq:correction}
\end{equation}
\textcolor{black}{Note that in both methods, ${\bf \tilde{x}}(t)$ is the augmented and improved state space. In the former approach, a discrepancy model of the missing physics is learned in the dynamical space, while in the latter approach, a discrepancy model for the systematic residual is constructed in the state space.} Equations~(\ref{eq:learning}) and (\ref{eq:correction}) are explicitly the focus of this manuscript.
Table~\ref{tbl:abrv} summarizes all the variable definitions whereas Fig.~\ref{fig:MainFigure} illustrates the discrepancy modeling framework.

By focusing explicitly on the discrepancy between measured and modeled dynamics, our framework shifts the view of discrepancies as 'errors' or 'residuals' to highly valuable measures for model improvement. The field of data-driven engineering for dynamical systems has an opportunity to improve system characterization by disambiguating deterministic and random effects within the model-measurement mismatch. In this paper, we formalize the mathematical infrastructure of discrepancy modeling for dynamical systems, highlighting the interplay and balance between deterministic and random effects. Specifically, we consider four data-driven modeling methods for identifying discrepancy models and provide two principled approaches for evaluating the utility and suitability of discrepancy modeling \textcolor{black}{in practical engineering applications}. We leverage recent mathematical advancements in data-driven model discovery and evaluate the interplay between \textcolor{black}{missing physics, $g({\bf x}(t))$, and random processes, ${\cal N}(\mu,\sigma)$,} showing how their relative sizes determine the ability to disambiguate deterministic from random or noisy effects. To encourage exploration and expansion of discrepancy modeling, we employ base coding packages for each model discovery method implementation~\cite{champion2019data, champion2020unified, rudy2017data,tu2013dynamic,rasmussen2003gaussian, narendra1992neural, brunton2016discovering, kutz2016dynamic, bagheri2014effects, askham2018variable, brunton2017chaos, de2020pysindy, swiler2020survey,kaheman2022automatic}. We found that the performance characteristics of our suite of model discovery methods matched their documented performance for dynamical systems modeling.  
All the methods proposed (SINDy, DMD, GPR \& NN) can be used successfully for both modeling the missing physics or the systematic residual.  Thus, the evaluation of an appropriate method involves the intent of the user (\textit{e.g.}, interpretability of the discrepancy model), \textcolor{black}{constraints of a practical engineering applications (\textit{e.g.}, if the Platonic model is accessible),} and the computational efficiency and robustness of the method for a given set of measurements (\textit{e.g.}, data quantity, quality, resolution).

\section{Methods for Data-Driven Modeling}

\subsection{General Forms}

\textcolor{black}{In this section we briefly present the data-driven modeling methods used for learning discrepancy models in their general forms. It should be noted that there are many variants and improvements that can be implemented for each method considered.} However, this comparative study does not aim to optimize the ability of each technique to perform at its absolute best.  Indeed, hyper-parameter tuning of each technique is problem specific and again depends upon measurement characteristics (quantity, quality, resolution) and the intent of its use.  Thus our goal is to demonstrate the various possibilities and their appropriate uses.  
{\color{black}It should be noted that our error metrics are generally prescribed by least-square fitting.  Although more sophisticated loss functions can be imposed to improve performance in the different modeling paradigms, alternative error metrics are not considered here.}

\subsection{Gaussian Process Regression (GPR) }
% What is GPR
{\em Gaussian process regression} (GPR) is a non-parametric supervised learning approach to regression that uses a Gaussian process prior for Bayesian inference~\cite{rasmussen2003gaussian}. Not only is GPR a powerful nonlinear interpolation tool, its inherent probabilistic architecture allows for uncertainty quantification of interpolated values. \textcolor{black}{Impressively, this method is a universal approximator with a closed form solution.}
%
% How does it work
Consider the dataset $D\{({\bf X}_i,{\bf Y}_i | i = 1,\dots, n\}$ which is split into test and training subsets where ${\bf X}_i$ are individual observations and ${\bf Y}_i$ are observation labels. A Gaussian process $f(\textbf{X}\textcolor{black}{_i})$ is defined by a mean function $m(\textbf{X}\textcolor{black}{_i})$ and covariance function $\textcolor{black}{\kappa} (\textbf{X}\textcolor{black}{_i},\textbf{X}\textcolor{black}{_j})$ as
\begin{eqnarray*}
    m(\textbf{X}\textcolor{black}{_i}) &=& \mathbb{E}[f(\textbf{X}\textcolor{black}{_i})] \\
    \textcolor{black}{\kappa} (\textbf{X}\textcolor{black}{_i},\textbf{X}\textcolor{black}{_j}) &=& \mathbb{E}[(f(\textbf{X}\textcolor{black}{_i}) - m(\textbf{X}\textcolor{black}{_i}))(f(\textbf{X}\textcolor{black}{_j}) - m(\textbf{X}\textcolor{black}{_j}))]
\end{eqnarray*}

such that the Gaussian process is:

\begin{align}
   f(\textbf{X}\textcolor{black}{_i}) \sim \mathcal{GP}(m(\textbf{X}\textcolor{black}{_i}), \textcolor{black}{\kappa} (\textbf{X}\textcolor{black}{_i},\textbf{X}\textcolor{black}{_j})).
\end{align}

Kernel functions are used to calculate covariance and enforce assumptions about the data. 
Specifically, kernels are parameterized by hyperparameters controlling covariance characteristics (\textit{e.g.} length-scales or periodicity) and are optimized by maximizing the log marginal likelihood during model selection. 
As the prior distribution is a Gaussian process $f(\textbf{X}\textcolor{black}{_i}) \sim \mathcal{GP}$, the conditional distribution $f(\textbf{X}\textcolor{black}{_i})| D_n$ is the posterior, i.e. the \textit{predictive distribution}. 
Computing on a finite-sized data set and partitioning it into \textcolor{black}{training inputs ($\bf X$), training outputs ($\bf f$), test inputs ($\bf X_*$), and test outputs  ($\bf f_*$)}, the prior joint distribution is assumed to take the form:
\begin{align}
    \begin{bmatrix}
         {\bf f} \\ {\bf  f}_*
    \end{bmatrix} 
    \sim
    \mathcal{N}
    \left(
    \mathbf{0},
    \begin{bmatrix}
    \textcolor{black}{\kappa}({\bf X},{\bf X}) & \textcolor{black}{\kappa}({\bf X},{\bf X}_*) \\
    \textcolor{black}{\kappa}({\bf X}_*,{\bf X}) & \textcolor{black}{\kappa}({\bf X}_*,{\bf X}_*).
    \end{bmatrix}
    \right)
    \label{eq:GPR}
\end{align} 
To find the posterior, or predictive, distribution, the joint prior distribution must be restricted, or conditioned, to contain functions that agree with the observations, $\mathbf{f_*} | \textcolor{black}{{\bf X_*}, {\bf X}}, \mathbf{f} \sim \mathcal{N}(\Bar{\mathbf{f_*}}, \text{cov}(\mathbf{f_*}))$, which can be written in closed form~\cite{rasmussen2003gaussian}. 
An important note: GPR for prediction or parameter estimation is computationally expensive. Calculating the maximum likelihood requires finding the determinant and inverse of the covariance matrix, which has cubic computational complexity.  Ultimately, GPR is about regressing to a Gaussian distribution and estimating the appropriate variances via (\ref{eq:GPR}).  The mathematical details of GPR can be found here~\cite{rasmussen2003gaussian}.

{\color{black}
GPR is the most general method advocated for discrepancy modeling.  It simply gives the best fit Gaussian distribution, with estimates for the mean and variance with minimal hyper-parameter tuning in the optimization.  However, its interpretability lies in ones choice of kernel function to describe characteristics of the discrepancy, such as periodicity or smoothness.  The algorithm is generalizable provided the data is drawn from a stationary distribution, which may not be the case in practice.  Instead it simply quantifies the statistics of the error.  It is especially useful when large noise, or small signal-to-noise ratios, are present in the system. It is recommended to use additional techniques when using large datasets to avoid costly computation due to covariance matrix inversion.}

% What is DMD
\subsection{Dynamic Mode Decomposition (DMD)}
{\em Dynamic mode decomposition} (DMD) is a modern system identification approach based on data-driven regression. DMD extracts spatio-temporal structures from time series data and learns a low-dimensional linear model describing the evolution of features that encode salient system behavior. 
Consider a dynamical system measured at evenly-spaced time points ${\bf t}=[t_1, t_2, \cdots, t_m]$. From measurements {\color{black}${\bf x}_k={\bf x}(t_k)$}, we construct a matrix of snapshots ${\bf X}({\bf t})$ = $[{\mathit {\bf x}_1 \,\, {\bf x}_2\,\,  \dots \,\, {\bf x}_n}]$ $\in \mathbb{R}^{m\times n}$. A discrete time linear representation of a system is assumed to take the standard form, ${\bf x}_{t+1} = {\bf A x}_t$ such that the linear operator $\bf A$ progresses the state vector ${\bf x}_t$ forward in time. Mirroring this standard form, two snapshot matrices are defined as:
\begin{equation}
    {\bf {X}} =
    \begin{bmatrix}
      | & | &  & | \\
      {\bf x}_1 & {\bf x}_2 & \cdots & {\bf x}_{m-1} \\
      | & | &  & | 
    \end{bmatrix},
  \label{eq:DMD_X1}
\end{equation}
\begin{equation}
    {\bf {X}}' =
    \begin{bmatrix}
      | & | &  & | & \\
      {\bf x}_2 & {\bf x}_3 & \cdots & {\bf x}_{m} & \\
      | & | &  & | &
    \end{bmatrix}
  \label{eq:DMD_X2}
\end{equation}
where ${\bf X}'$ is the time-shifted matrix of snapshots ${\bf X}$, \textit{i.e.} ${\bf X}' = {\bf AX}$. The \textit{exact} DMD is the best fit linear mapping ${\bf {A}}$ between snapshot pairs $\bf X$ and ${\bf X}'$:
\begin{equation}
    {\bf {A}} = \argmin_{\bf A} \| {\bf X' - AX} \| _{F} = {\bf X}'{\bf X}^{\dagger}
  \label{eq:pseudoinverse}
\end{equation}
where $\| \cdot \|_{F}$ is the Frobenius norm and $\dagger$ is the Moore-Penrose pseudoinverse~\cite{tu2013dynamic}. DMD exploits the singular value decomposition (SVD) to solve for:
\begin{equation}
    {\bf {A}} = {\bf U}^*{\bf X}'{{\bf V}}{\bf \Sigma}^{-1}.
  \label{eq:A}
\end{equation}
DMD exploits low-rank structure of high-dimensional systems, and thus projects $\bf A$ onto the first \textit{r} modes of the principle components $\bf{U}_r$. This rank-\textit{r} truncation of $\bf{X} \approx \bf{U}_r\bf{\Sigma}_r\bf{V}^*_r$ approximates the pseudo-inverse:
\begin{equation}
    {\bf \Tilde{A}} = {\bf U}^*_r{\bf A}{\bf U}_r = {\bf U}^*_r{\bf X'}{\bf V}_r{\bf \Sigma}^{-1}_r .
  \label{eq:A_hat}
\end{equation}
The eigendecomposition of $\Tilde{\bf A}$ yields the eigenvalues and eigenvectors, $\bf \Tilde{A}W = W\Lambda$, which provides insight into underlying system properties such as growth modes and resonance frequencies ~\cite{tu2013dynamic}. 
The eigenvectors of $\bf A$ are the \textit{DMD modes} $\bf \Phi$:
\begin{equation}
    {\bf \Phi} = \bf{X'}\tilde{\bf V}\tilde{\Sigma}^{-1}\bf{W}.
  \label{eq:eigA}
\end{equation}
Along with the mode amplitudes, ${\bf b} = {\bf \Phi}^{\dagger}{\bf x}_1$, the well known DMD solution takes the form
\begin{equation}
    {\bf x}(t) = \sum_{i = 1}^r {\phi}_i e^{\omega_i t} {b}_i = \bf{\Phi} \text{exp}({\bf \Omega}t)\bf{b}.
  \label{eq:DMD_solution}
\end{equation}
However, exact DMD is prone to biased errors resulting from noisy measurements, affecting model fit and forecasting stability~\cite{bagheri2014effects}. Therefore, Askham and Kutz~\cite{askham2018variable} introduced \textit{optimized} DMD, which uses variable projection to perform nonlinear optimization for de-biasing model fitting in the presence of observation noise. More specifically, the variable projection method optimally computes nonlinear least squares exponential fitting for DMD:
\begin{equation}
    \argmin_{\omega, \Phi_b} \| {\bf X - \Phi \text{exp}(\Omega}t){\bf b}  \| _{F}.
\end{equation}
In this paper, we use optimized DMD, but generally refer to it as 'DMD'. 
Note: the rank of $\bf A$ cannot exceed the state dimension of $\bf X$, and DMD algorithms rely on the availability of full-state measurements, typically of high dimensions. When only partial observations or low-dimensional system measurements are available, it is helpful to build an augmented state vector that is 'lifted' into a higher dimension. For our study, we accomplish this via time-delay embedding, which also happens to result in an intrinsic coordinate system forming a Koopman-invariant subspace in which nonlinear dynamics appear linear~\cite{brunton2017chaos}. 

{\color{black}
The DMD discrepancy architecture provides a potentially interpretable model for characterizing the dynamics, allowing for the decomposition of the data into modes and frequencies.  DMD is quite robust, especially in its newest versions such as the {\em bagging optimized DMD} (BOP-DMD)~\cite{sashidhar2022bagging} which leverages statistical bagging to help stabilize the linear model while providing uncertainty quantification.  It scales well and has been shown to have some degree of generalizability, with minimal hyper-parameter tuning.  However, it is so efficient to compute that one can easily compute new DMD models on the fly.  BOP-DMD can always be computed, making it an attractive method as an alternative to GPR, handling even larger noise fluctuations with statistical bagging.}

\subsection{Sparse identification of nonlinear dynamics (SINDy)}

% What is SINDy
{\em Sparse identification of nonlinear dynamics} (SINDy) recovers parsimonious representations of the dynamics from measurement data by sparse regression to a library of candidate models~\cite{brunton2016discovering,champion2019data,rudy2017data}.
% How does it work
Consider a nonlinear dynamical system measured at time points ${\bf t}=[t_1, t_2, \cdots, t_m]$. From measurements, we construct the matrix ${\bf X}({\bf t})$ = $[{\mathit {\bf x}_1({\bf t}) \,\, {\bf x}_2({\bf t})\,\,  \dots \,\, {\bf x}_n({\bf t})}]$ $\in \mathbb{R}^{m\times n}$. The method introduced in \cite{brunton2016discovering} seeks to identify ${\bf f}$ via sequential threshold least-squares, which is a proxy for the sparsifying zero-norm. The set of ${\mathit n}$ state measurements are used to populate a library of candidate nonlinear terms ${\bf \Theta (X) = [1^\top \; X^\top \; (X \otimes X)^\top \; \cdots \; \text{sin}(X)^\top ]}$, where ${\bf x \otimes y}$ defines the vector of all product combinations of the state components. Each candidate term should be unique, as a suitable library is crucial in the SINDy algorithm. A common strategy is to start with polynomials and increase the complexity of the library with other terms, such as trigonometric functions. Thus, a dynamical system can be re-written as:
\begin{equation}
    {\bf \dot{X} = \Theta(X)\Xi } .
  \label{eq:SINDy}
\end{equation}
The time derivatives ${\bf \dot{X}}({\bf t}) = [{\mathit {\bf \dot{x}}_1({\bf t}) \,\, {\bf \dot{x}}_2({\bf t}) \,\, \dots \,\, {\bf \dot{x}}_n({\bf t})}]$, if not measured directly, can be found via numerical differentiation and should be appropriately de-noised, if necessary \textcolor{black}{~\cite{van2020numerical, gottwald2021supervised, chen2022autodifferentiable, farchi2021using,chartrand2011numerical,cullum1971numerical,ramm2001stable}}. The coefficients ${\bf \Xi}$ are the {\em sparse} weightings of the corresponding candidate library terms. Therefore, our regression relies on sparse regularization to enforce a parsimonious $\bf \Xi$ corresponding to the fewest nonlinear terms in our library that describe our dynamics well: 
\begin{equation}
    {\bf \Xi} = \text{arg}\,\min\limits_{\hat{ \bf \Xi}\, } \Vert{\bf \Theta(X)\hat{\bf \Xi} - \dot{X}}\Vert_\text{2} + \lambda\Vert{\bf \hat{\bf \Xi}}\Vert_\text{0} 
  \label{eq:sparse}
\end{equation}
Regressing to the zero-norm is often achieved by relaxing the one-norm.  However, modern optimization frameworks are allowing for computationally tractable proxies for the zero-norm that are superior to the one-norm relaxation~\cite{champion2020unified}.

{\color{black}
The SINDy algorithm has the strongest potential in providing an interpretable model for discovering missing physics. However, this method is not expected to perform well in fitting a discrepancy model to the state-space residual, as residuals might not be amenable to a sparse representation. SINDy also has the best possibility for generalization since the model learned is minimally parameterized.  The algorithm is also computationally efficient and scalable, with minimal hyper-parameter tuning.  Like BOP-DMD~\cite{sashidhar2022bagging}, ensemble SINDy~\cite{fasel2021ensemble} has been developed to help make SINDy robust even with increasing noise.  Even with such methods for improving performance, SINDy is not as robust as GPR and DMD in handling noise.  Thus it is recommended for low and intermediate noise regimes.}

\subsection{Neural Networks (NN)}
Artificial neural networks, or simply \textit{neural networks} (NN), are mathematical models inspired by biological neural networks. While there is a wide array of literature on NN~\cite{narendra1992neural,gonzalez1998identification,krischer1993model,rico1994continuous,lagaris1998artificial, daw2017physics,lu2019deeponet, raissi2019physics}, we outline the basic concepts. NN learn a mapping between a set of input data and target outcomes. The middle, or hidden, layers form a compositional structure that optimizes the association between the training data set. The user designates the depth (number of hidden layers), the dimensionality (number of nodes) of each layer, and how each layer is connected. 

Linearly, this means a NN optimizes over the compositional function to learn the neural network weights and biases matrices ${\bf A}_j$ between the $k$-hidden layers: 
\begin{equation}
    \argmin_{{\bf A_\textit{j}}} (f_M({\bf A}_M , \dots , f_2({\bf A}_2, f_1 ({\bf A}_1,{\bf x})) \dots) + \lambda g({\bf A}_j)),
    \label{eq:NN_general}
\end{equation}
where $\lambda g({\bf A}_j)$ is included to provide and appropriate regularization for the solution. For example, a simple, single hidden layer $(k = 1)$ NN is structured as:
\begin{equation}
\begin{split}
    {\bf x}^{(1)} &= {\bf A}_1 {\bf x} \\
    {\bf y} &= {\bf A}_2 {\bf x}^{(1)} \\
\end{split}.
\end{equation}
Leveraging the compositional structure, a mapping is defined by:
\begin{equation}
    {\bf y} = {\bf A}_2{\bf A}_1 {\bf x},
\end{equation}
which generalizes to $M$ layers 
\begin{equation}
    {\bf y} = {\bf A}_M{\bf A}_{M-1} \dots {\bf A}_2 {\bf A}_1 {\bf x}.
\end{equation}
Nonlinear mappings are structured similarly. In this case, nonlinear activation functions connect hidden layers and are given by:
\begin{equation}
\begin{split}
    {\bf x}^{(1)} &= f_1 ( {\bf A}_1, {\bf x}) \\
    {\bf y} &= f_2 ( {\bf A}_2, {\bf x}^{(1)}). \\
\end{split}
\end{equation}
Further, nonlinear activation functions, $f_j(\cdot)$, can differ between layers. Thus, nonlinear mapping between a given set of input and output data over $M$ layers is structured as:
\begin{equation}
    {\bf y} = f_M({\bf A}_M , \dots , f_2({\bf A}_2, f_1 ({\bf A}_1,{\bf x})) \dots) = {\bf f}_{\boldsymbol{\theta}}({\bf x})
\end{equation}
where ${\bf f}_{\boldsymbol{\theta}}(\cdot)$ represents the overall network structure with weights and biases $\boldsymbol{\theta}$.
While often used for classification, NNs can be structured to learn the evolution of dynamical systems. NN for dynamical systems provides a flexible and powerful architecture for high-dimensional supervised learning of system behavior for future state predictions~\cite{brunton2019data}.
%For example, for a system ${\bf x}_{t+\Delta t} = f({\bf x},t)$, if the input layer contains state measurements at time $t$ and the output layer contains state measurements at time $t + \Delta t$, DL will learn the best-fit model parameters to map the system's spatio-temporal behavior. Similarly (yet distinctly), for a system $\frac{d{\bf x}}{dt} = f({\bf x},t)$, if the input data contains state measurements at time $t$ and the the output layer contains the state derivatives $\frac{d{\bf x}}{dt}$, DL learns the best-fit activation function parameters that describe the state-space dynamical evolution. 

%\subsection{Factors Influencing Discrepancy Disambiguation}
% Should be 2.1?
%What influences disambiguation / model fitting / recover dynamics : data quality (noisy data sets), quantity (short data sets), structure (model fit); 

%External factors must be considered when evaluating the utility and suitability of discrepancy modeling. Here, we consider (1) data quantity, (2) data quality, and (3) discrepancy structure. [REF]

%DISCUSS EXPECTED SUCCESS/FAILURE IN DDM?

{\color{black}
Of the methods advocated, neural networks are the least interpretable, providing a black-box algorithm that is trained from data.  The quality of the models trained are also sensitive to noise, making it more delicate than some of the other algorithms.  However, as with many NN applications, with high-quality data of sufficiently large volume, the NN can provide high-quality discrepancy models in practice.  It is the most computationally expensive of all the algorithms, with computational costs greatly exceeding other algorithms and requires significant hyper-parameter tuning in general. \textcolor{black}{Techniques such as stochastic gradient descent greatly improve network optimization.} Neural networks are recommended when large quantities of data of low and intermediate noise are available.}

\section{Discrepancy Modeling Framework}

\subsection{The Two Approaches}

{Discrepancy modeling} aims to improve system characterization from data by disambiguating deterministic and random effects within the model-measurement mismatch. There are two nuanced, yet distinct, means of building a discrepancy model of missing physics by learning a model of the (i) deterministic dynamical error, or (ii) systematic state-space residual. In the first approach, the discrepancy is learned in the dynamical space, \textit{i.e.}, the discrepancy model learns the error between the derivative of the measurements, $\dot{{\bf y}}_k$, and the approximate dynamics given the state measurements, $f({\bf y}_k)$. Thus, learning a discrepancy model is akin to learning missing physics and is to be appended to the approximate dynamics. In the second approach, the discrepancy is learned in the state space, \textit{i.e.}, the discrepancy model learns the systematic residual between the approximate state space, ${\bf x}(t_k)$, and the measurements, ${\bf y}_k$. Thus, this discrepancy model acts as a correction to the approximate state-space solution. 
\subsection{Learning the Deterministic Dynamical Error}

In this approach, the discrepancy model learns the deterministic dynamical error within the model-measurement mismatch to improve the known, approximate dynamical model. The discrepancy is learned in the dynamical space to resolve the error between measurement derivatives and approximate dynamics, ${\textcolor{black}{\bf \dot{x}}_D}(t_k) = \dot{\bf y}_k  - f({\bf y}_k) $. 
\textcolor{black}{
The formulation begins with Eqns. (\ref{eq:plato}-\ref{eq:measurements}). The discrepancy model $\color{black}{{\delta}_f}(\cdot)$ estimates the dynamical relationship between measurements and deterministic dynamical error:
\begin{equation*}
        {\bf \dot{x}_D}(t_k) \approx {\delta}_f({\bf y}_k)
        \label{eq:discrepancy_dynamical}
\end{equation*}
}
Using the suite of data-driven methods proposed, we can build discrepancy models $\color{black}{{\delta}_f}(\cdot)$:
\begin{equation*}
\begin{split}
    \text{GPR}: \quad & {\textcolor{black}{\bf \dot{x}}_D(t_k)} \sim \mathcal{GP}(m({\bf y}_k), \textcolor{black}{\kappa} ({\bf y}_k,{\bf y}_{k+1})) \\
    \text{DMD}: \quad & {\textcolor{black}{\bf \dot{x}}_D(t_k)} \approx {\boldsymbol \Phi} \; \text{diag}({\bf b}) e^{{\boldsymbol \omega}{t_k}}, \quad {\bf b} = {\boldsymbol \Phi}^{\dagger} {\bf y}_1 \\
    \text{SINDy}: \quad & {\textcolor{black}{\bf \dot{x}}_D(t_k)} = \Theta({\bf y}_k)\Xi \\
    \text{NN}: \quad & {\textcolor{black}{\bf \dot{x}}_D(t_k)} = {f}_{\boldsymbol{\textcolor{black}{\theta}}}({\bf y}_k)
\end{split}
\end{equation*}
and appended it to the approximate dynamics:
\begin{equation*}
        {\dot{\tilde{\bf x}}}(t) = \tilde{\mathit F}(\tilde{\bf x}(t)) = \mathit{f}(\tilde{\bf x}(t)) + {\textcolor{black}{{\delta}\mathit{_f}}}(\tilde{\bf x}(t)) 
        \label{eq:discrepancy_dynamical_model}
\end{equation*}
to minimize:
\begin{equation}
       {\bf \dot y}_k - \tilde{F}({\bf y}_k) 
       \label{eq:aug_dyn_error}
\end{equation}

\subsection{Modeling the Systematic State-Space Residual}
In this approach, the discrepancy model learns the time evolution of systematic state-space residual and corrects the approximate state-space solution. The discrepancy is learned in the state space to resolve the residual between measurements and the approximate solution, ${\textcolor{black}{{\bf x}_D}(t_k)} = {\bf y}_k - {\bf x}(t_k)$. The discrepancy model framework begins with Eqs. (\ref{eq:plato}-\ref{eq:measurements}). 
\textcolor{black}{The discrepancy model $\color{black}{{\delta}}(\cdot)$ estimates the state-space relationship between the approximate state space and the systematic residual:
\begin{equation*}
        {\bf {x}_D}(t_k) \approx {\delta}({\bf x}(t_k))
        \label{eq:discrepancy_residual}
\end{equation*}
}
Using the suite of data-driven methods proposed, we can build discrepancy models $\color{black}{{\delta}}(\cdot)$:
\begin{equation*}
\begin{split}
    \text{GPR}: \quad &  {\textcolor{black}{\bf {x}}_D(t_k)} \sim \mathcal{GP}(m({\bf x}(t_k)), \textcolor{black}{\kappa} ({\bf x}(t_k),{\bf x}(t_{k+1})) \\
    \text{DMD}: \quad & {\textcolor{black}{\bf {x}}_D(t_k)} \approx {\boldsymbol \Phi} \; \text{diag}({\bf b}) e^{{\boldsymbol \omega}{t_k}}, \quad {\bf b} = {\boldsymbol \Phi}^{\dagger} {\bf x}(t_1) \\
    \text{SINDy}: \quad & {\textcolor{black}{\bf {x}}_D(t_k)} = \Theta({\bf x}(t_k))\Xi \\
    \text{NN}: \quad & {\textcolor{black}{\bf {x}}_D(t_k)} = {f}_{\boldsymbol{\textcolor{black}{\theta}}}({\bf x}(t_k))
\end{split}
\end{equation*}
and correct the approximate state-space solution (\textcolor{black}{initialized using the first observation ${\bf y}_1$}):
\begin{equation*}
       {\bf \tilde{x}}(t) = {\bf x}(t) + {\textcolor{black}{{\delta}}}({\bf x}(t))
       \label{eq:discrepancy_residual_model}
\end{equation*}
to minimize:
\begin{equation*}
       {\bf y}_k -\tilde{\bf x}(t_k).
\end{equation*}
\section{Discrepancy Modeling Applications}
Discrepancy modeling is both system- and situational-dependent; thus we evaluate the utility and suitability of each discrepancy modeling approach using a suite of four model discovery methods by comparing reconstruction and forecasting accuracy. We further probe discrepancy modeling performance on increasingly complex systems and for increasing levels of noise. 
\subsection{Van der Pol Oscillator: A Simple Example}

\begin{figure}[t]
    \centering
    \includegraphics{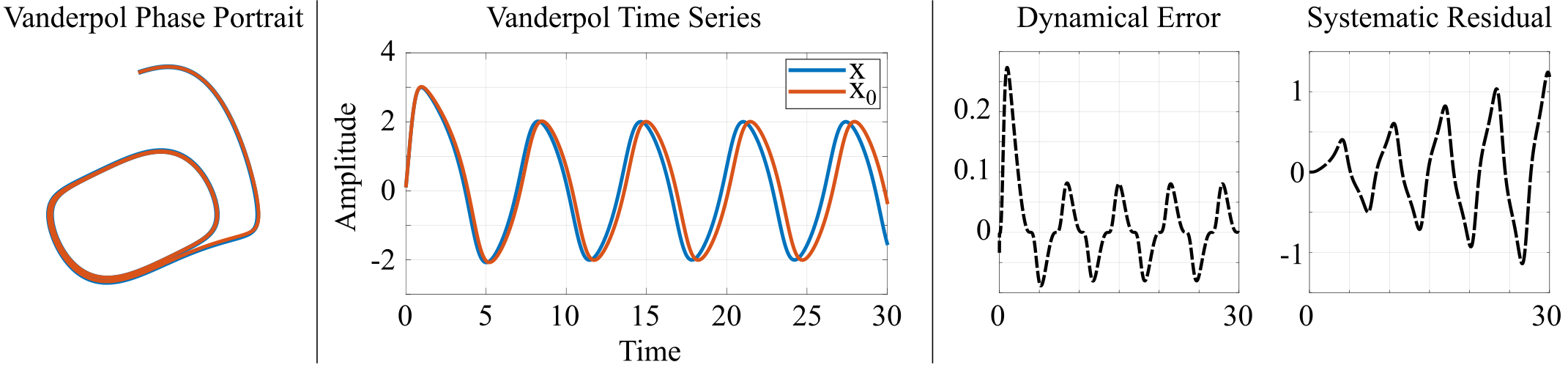}
    \caption{Van der Pol oscillator example (no noise) with and without a discrepancy. While the salient dynamical features are preserved as seen in the phase portrait (left panel), the time evolution (middle panel) diverges quickly with only an $\epsilon$-small dynamical difference. The dynamical error and systematic residual (right panel) are plotted to demonstrate the two discrepancy types.}
    \label{fig:Vanderpol}
\end{figure}

We began with a simple model and no noise. The data used in this example was generated using the Van der Pol oscillator:
\begin{equation}
    \frac{d^2x}{dt^2} - \mu (1-x^2)\frac{dx}{dt} + x = 0
    \label{eq:Vanderpol_platonic}
\end{equation}
 using \textcolor{black}{$x_{0}(0) = x(0) = [0.1, 5]$}, $t = [0, 50]$, and $\Delta t = 0.01$. We simulated Eqn.~(\ref{eq:Vanderpol_platonic}) to generate our Platonic or approximate dynamics. To this system, we added a small nonlinear term $\epsilon x^3$:
\textcolor{black}{
\begin{equation}
    \frac{d^2x_{0}}{dt^2} - \mu (1-x_{0}^2)\frac{dx_{0}}{dt} + x_{0} + \epsilon x_{0}^3 = 0
    \label{eq:Vanderpol_truth}
\end{equation}
}
Eqn.~(\ref{eq:Vanderpol_truth}) was simulated to generate the true system behavior using $\epsilon = 0.01$. This $\epsilon$-small nonlinearity represents the missing deterministic physics not captured in the approximate model. This $\epsilon$ cubic term added to the approximate dynamics perturbed the time evolution of Van der Pol, as seen in Fig.~(\ref{fig:Vanderpol}), while still maintaining salient characteristics associated with the oscillator. 

\subsubsection{Deterministic Dynamical Error}
\begin{figure}[t]
    \centering
    \includegraphics[width=\textwidth]{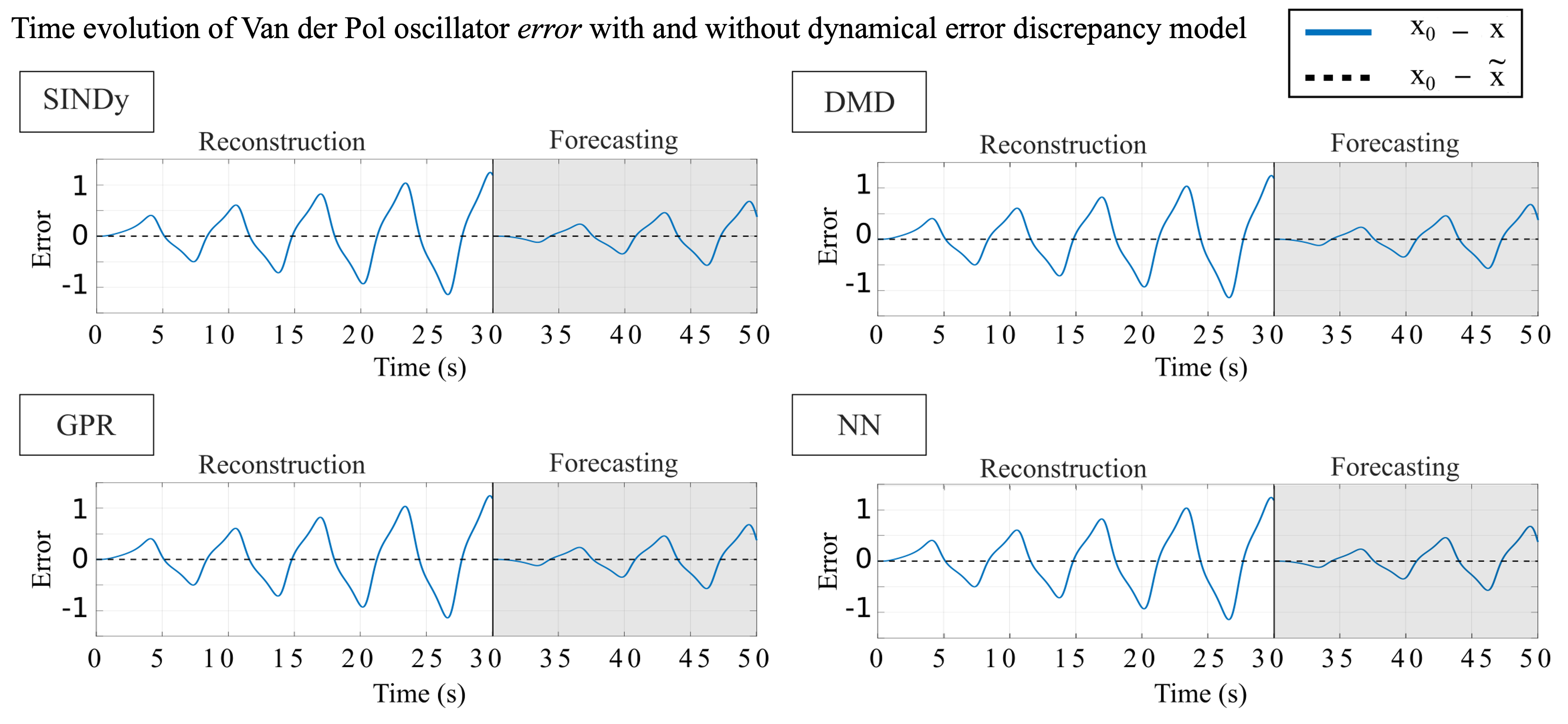}
    \caption{Remaining state-space error with and without a discrepancy model of the deterministic dynamical error appended to the approximate Van der Pol oscillator model (no noise). The blue line shows the error without a discrepancy model, and the black dashed line shows the error with a discrepancy model recovering the missing physics. The suite of model discovery methods learned the missing physics within the discrepancy such that no error remained.}
    \label{fig:Vanderpol_IDphysics}
\end{figure}

We first evaluated the ability of discrepancy modeling to recover missing physics for the Van der Pol oscillator. As seen in Fig~(\ref{fig:Vanderpol_IDphysics}), the suite of model discovery methods learned the deterministic dynamical error within the mismatch such that no error remained. Both the reconstruction and the forecasting errors (black dashed line) between the true and augmented models is zero for all model discovery methods. The discrepancy dynamics between the true and approximate models are denoted by the blue line. The Van der Pol oscillator has a parameter dictating the nonlinearity of its oscillations; we chose a mild level of nonlinearity to start. As the nonlinearity parameter increases, we would expect that \textit{linear} model discovery methods (\textit{e.g.}, DMD) would struggle to build an accurate discrepancy model and thus fail to fully recover missing physics.
\begin{figure}[t]
    \centering
    \includegraphics[width=\textwidth]{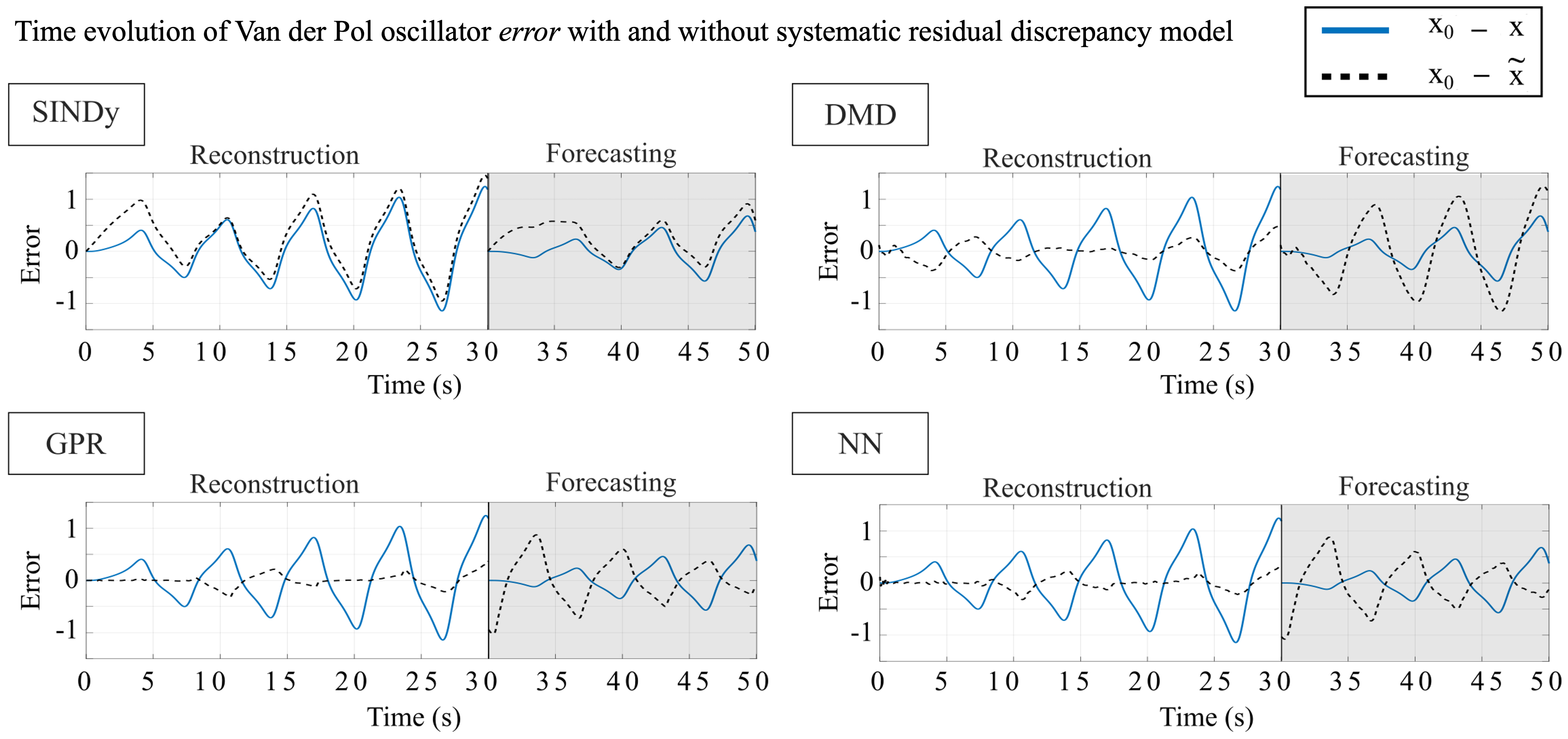}
    \caption{Remaining state-space error with and without a discrepancy model of the systematic state-space residual to correct the approximate Van der Pol oscillator state-space solution (no noise). The blue line shows the error without a discrepancy model, and the black dashed line shows the error with a discrepancy model modeling the systematic residual. The suite of model discovery methods had various success in modeling the systematic residual, resulting in promising utility of discrepancy modeling for correcting state-space solutions.}
    \label{fig:Vanderpol_errorModel}
\end{figure}
\subsubsection{Systematic State-Space Residual}
We next evaluated the ability of discrepancy modeling to learn the evolution of the systematic state-space residual for the Van der Pol oscillator. As seen in Fig.~(\ref{fig:Vanderpol_errorModel}), the suite of model discovery methods had various success in modeling the time evolution of the systematic residual, thus demonstrating promising utility of discrepancy modeling for correcting state-space solutions. SINDy failed to learn the systematic residual, thus resulting in reconstruction and forecasting errors similar to if no discrepancy model was used. The inability for SINDy to model systematic residual is unsurprising, as it is formulated to recover dynamical terms (See Eqn.~\ref{eq:SINDy}). DMD, GPR, and NN fared well in modeling the systematic residual, resulting in reduced error as compared to the approximate model; however, these methods struggled to correct forecasts of the approximate model. DMD was able to capture the salient features of the residual, yet the augmented solution appeared to be slightly time-shifted from the true solution, as seen by the brief zero-error within the reconstruction regime. Both GPR and NN learn promising discrepancy models of the residual, as they both had zero error between the true and augmented solutions for the first ten seconds in the reconstruction region, but began to diverge, albeit minimally. However, both GPR and NN had large errors at the beginning of forecasting. We theorize this is due to a failure to appropriately initialize. Generally, both methods learn their training trajectories with high accuracy; often, large datasets with many trajectories, and thus many initial conditions, were used in model training. In our case, because only one training trajectory was used, \textit{i.e.,} one set of initial conditions, GPR and NN were unable to extrapolate the expected systematic residual when initialized with a different initial condition for forecasting. We theorize this initialization error can be resolved to greatly reduce remaining error when using a state-space residual discrepancy model via: (i) increasing data \textit{quantity} (number of trajectories, initial conditions, parameters, resolution) for training or (ii) sacrificing the first portion of the test data/new times series to update the trained model.

\begin{figure}[t]
    \centering
    \includegraphics[width=\textwidth]{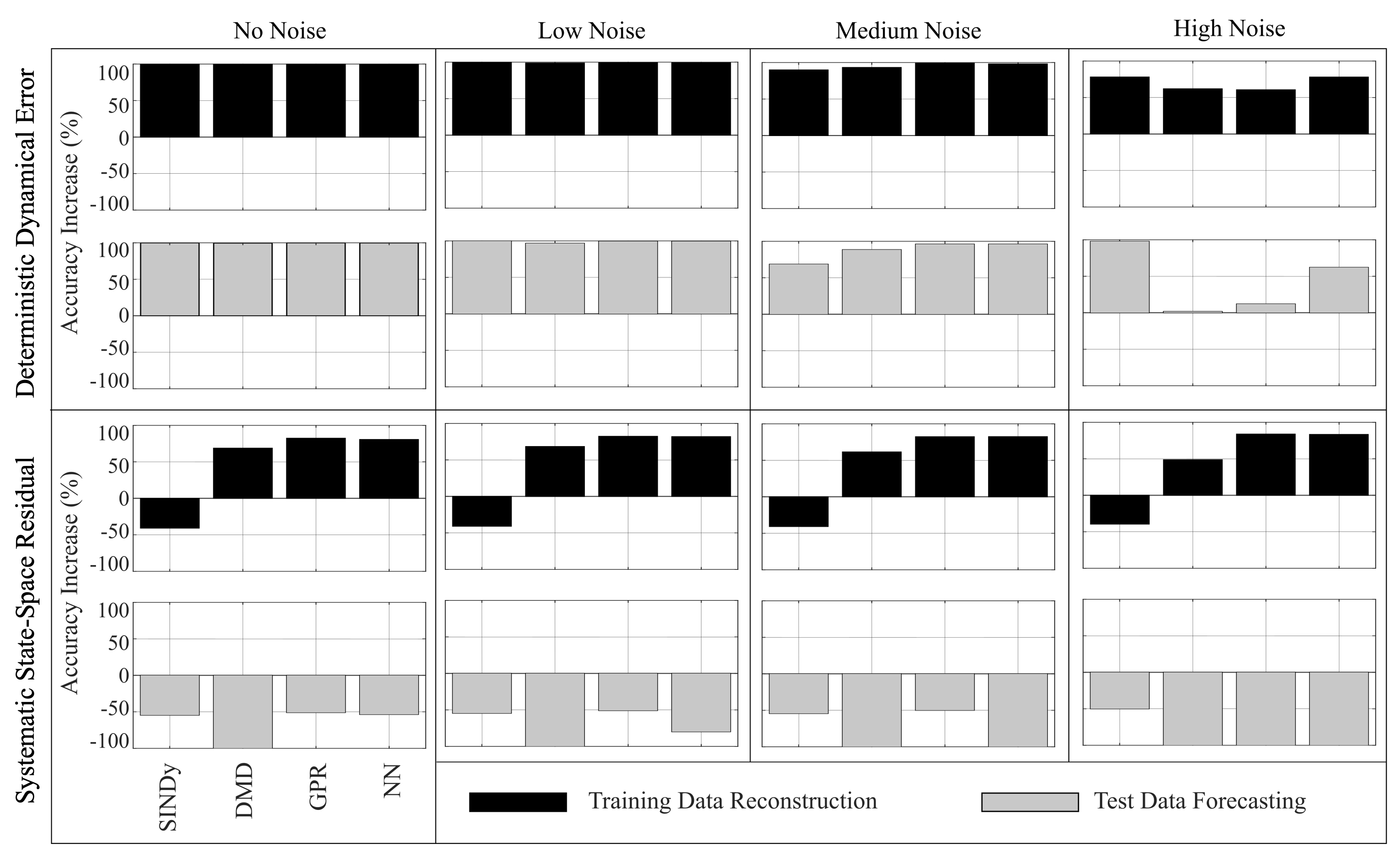}
    \caption{Increase in accuracy with discrepancy modeling augmentation in Van der Pol oscillator. Increase in \% accuracy is calculated as the relative change between the root mean squared errors (RMSE) of the augmented and approximate solutions as compared to the true solution. Results are shown for no (0\%), low (0.1\%), medium (1\%), and high (10\%) noise levels. Discrepancy modeling can be successful in numerous ways and depends on user intent (\textit{e.g.}, interpretability), data quality (\textit{e.g.}, noise tolerance), and data quantity (\textit{e.g.}, rapid evaluation vs. computational cost).}
    \label{fig:Vanderpol_RMSE_Compare}
\end{figure}

\subsubsection{Adding Noise}
\begin{figure}[t]
    \centering
    \includegraphics[width=\textwidth]{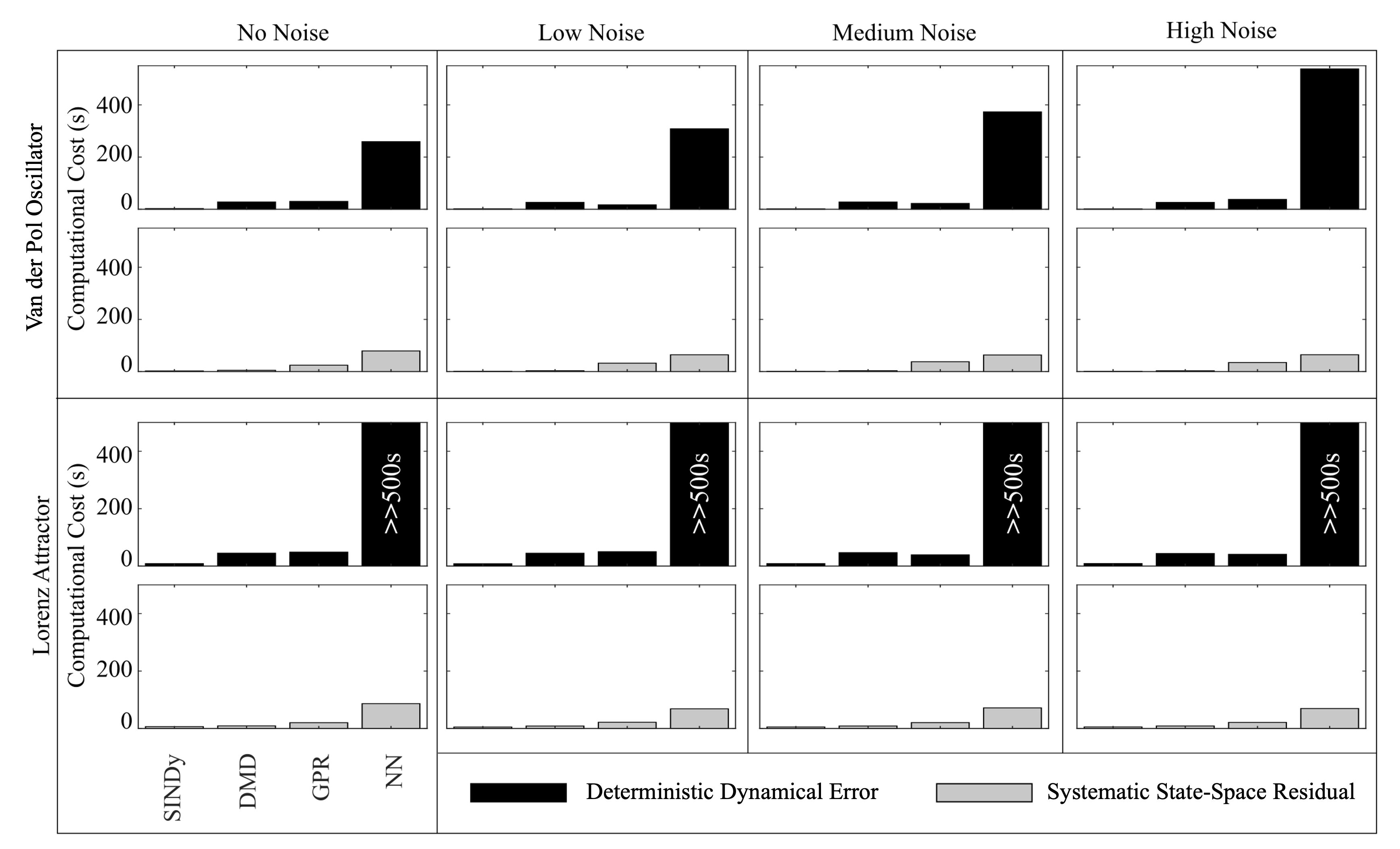}
    \caption{Computational costs (seconds) corresponding to executing the discrepancy modeling framework in MATLAB. Comparison across both approaches for each of the suite of model discovery methods. \textcolor{black}{Costs include data generation for approximate and true models, computation of discrepancy dynamics, and reconstruction and forecasting of augmented model.} The computational cost for the Van der Pol oscillator increases as noise increases for all model discover methods. SINDy has the lowest computational cost, followed by DMD, GPR, and NN, for both discrepancy modeling approaches. Learning missing physics with a NN has a noteably higher computational cost. This occurs because of how the discrepancy model is appended to the approximate dynamical model; the discrepancy dynamics are computed at each time step in MATLAB's ODE45 function, which greatly increases computational cost. These trends hold for the Lorenz attractor. Note: the time to execute the 'learning deterministic dynamical error' script using the NN was much higher than 500 seconds, and peaked around 2100 seconds.}
    \label{fig:Computational_Cost}
\end{figure}

We re-ran both discrepancy modeling approaches with the Van der Pol oscillator and increasing levels of Gaussian noise added to the 'true' Van der Pol observations to evaluate the effect of sensor measurement \textit{quality}. We use noise level of $\sigma = [0.1\%, \; 1\%, \; 10\%] $. As seen in Fig.~\ref{fig:Vanderpol_RMSE_Compare}, discrepancy modeling can be successful in numerous ways. Therefore, implementation amounts to user intent for discrepancy modeling, as well as the quantity and quality of sensor measurements. For example, when using discrepancy modeling to learn missing physics of the Van der Pol oscillator, all methods reconstruct and forecast true dynamics successfully in the no, low, and medium noise regimes. Note in the high noise regime, accuracy will most likely increase for methods like SINDy (which has an accuracy increase but is non-sparse) and DMD with innovations from baseline code packages (\textit{e.g.}, ensemble, bagging, culling)~\cite{de2020pysindy,sashidhar2022bagging,fasel2021ensemble}. If the user's intent for discrepancy modeling is interpretability, for no/low/medium noise, SINDy and DMD perform well, with minimal data quantity requirements and small computational costs (See Fig.~\ref{fig:Computational_Cost} for computational cost comparisons). Even if the user has no interest in interpretability, SINDy and DMD perform comparably to GPR and NN (methods that have increased data quantity requirements and greater computational costs). On the other hand, if the user's intent is to only reduce state-space error for example, correcting the state-space solution with a discrepancy model of the systematic residual is feasible with DMD, GPR, and NN. While forecasting errors appear abysmal, utility of these methods for forecasting may increase with better initialization, as discussed above. 

\subsection{Lorenz Attractor: Adding Chaos}
\begin{figure}[t]
    \centering
    \includegraphics{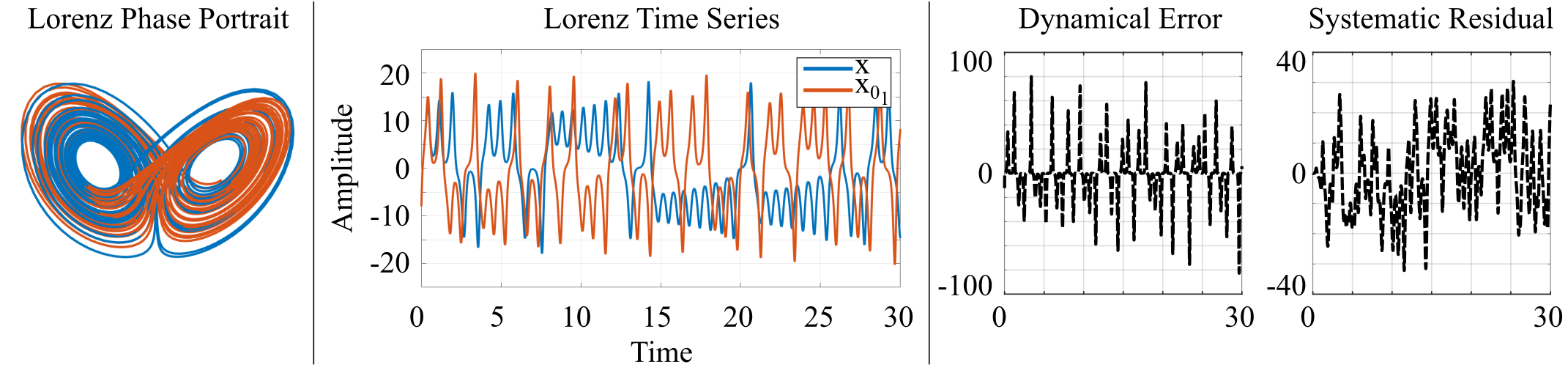}
    \caption{Lorenz attractor example (no noise) with and without a discrepancy. While the salient dynamical features are preserved as seen in the phase portrait (left panel), the time evolution (middle panel) bifurcates quickly with only an $\epsilon$-small dynamical difference. Chaotic dynamical systems like the Lorenz attractor are particularly sensitive to small errors in system dynamics. The deterministic dynamical error and systematic state-space residual (right panel) are plotted to demonstrate the two types of discrepancies.}
    \label{fig:Lorenz_System_Dynamics}
\end{figure}

We extended our discrepancy modeling analyses to a more complex canonical system. We simulated the Lorenz attractor, which has increasingly complex dynamical features like chaos:
\textcolor{black}{
\begin{equation}
%\begin{split}
    \frac{dx_1}{dt} = \sigma(x_2-x_1) \qquad
    \frac{dx_2}{dt} = x_1(\rho-x_3)-x_2 \qquad
    \frac{dx_3}{dt} = x_1x_2 - \beta x_3 
    \label{eq:Lorenz_platonic}
%\end{split}
\end{equation}
}
with the oft-used parameters, $\textcolor{black}{x_{0}(0) = x(0) = [-8; 8; 27],}  \sigma = 10$, $\rho = 28$, and $\beta = 8/3$, as well as $t = [0, 50]$ and $\Delta t = 0.01$, to generate our Platonic or approximate dynamics. To this system, we again added an $\epsilon$-small nonlinearity:
\textcolor{black}{
\begin{equation}
%\begin{split}
    \frac{dx_{0_1}}{dt} = \sigma(x_{0_2}-x_{0_1}) + \epsilon x_{0_1}^3   \qquad
    \frac{dx_{0_2}}{dt} = x_{0_1}(\rho-x_{0_3})-x_{0_2} \qquad
    \frac{dx_{0_3}}{dt} = x_{0_1}x_{0_2} - \beta x_{0_3} 
    \label{eq:Lorenz_truth}
%\end{split}
\end{equation}
}
and simulated to generate the true system behavior using $\epsilon = 0.01$. This $\epsilon$ cubic term added to the approximate dynamics perturbed the time evolution of Lorenz, as seen in Fig.~\ref{fig:Lorenz_System_Dynamics}, while still exhibiting salient characteristics associated with the attractor and maintaining dynamical stability.

\begin{figure}[t]
    \centering
    \includegraphics[width=0.95\textwidth]{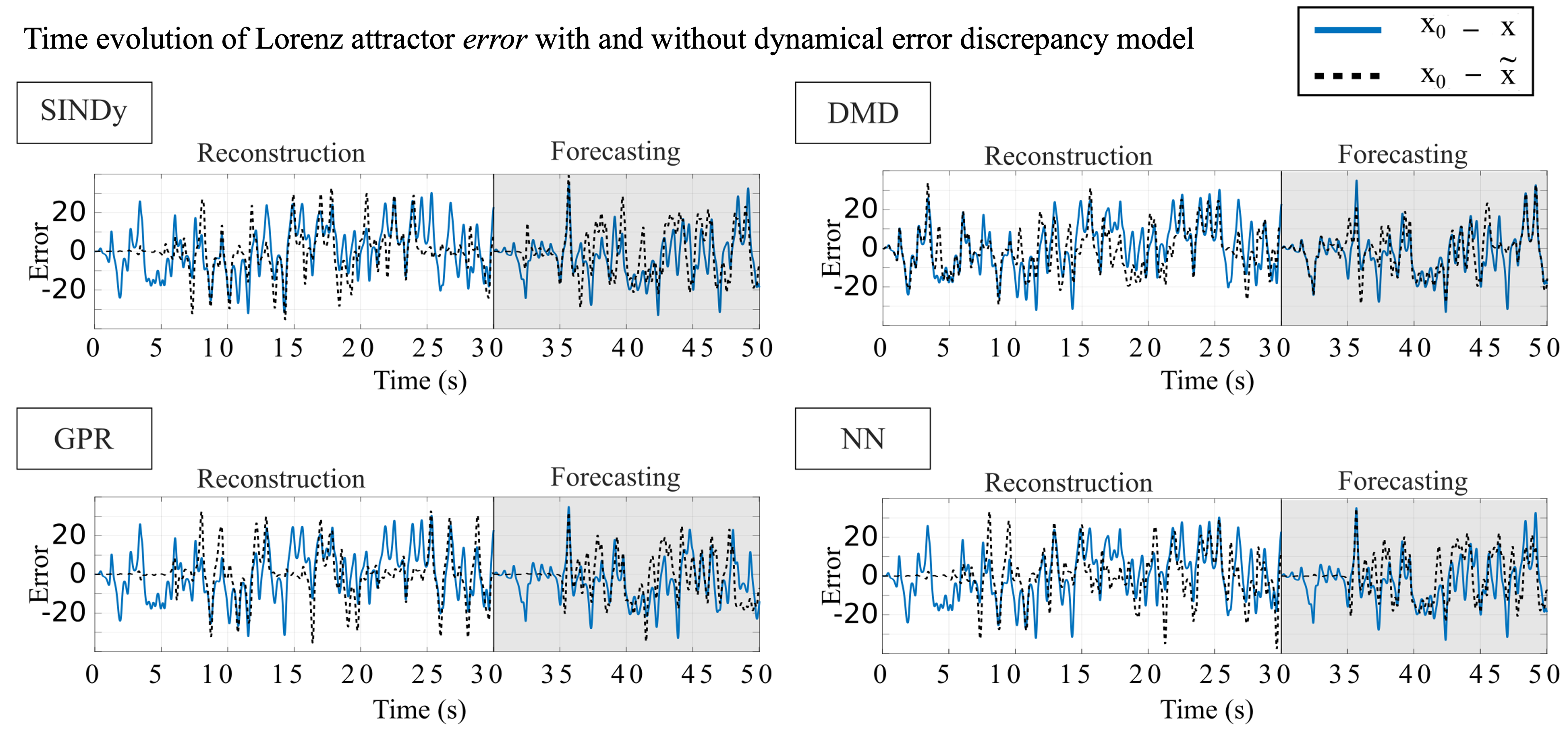}
    \caption{Remaining state-space error with and without a discrepancy model of the deterministic dynamical error appended to the approximate Lorenz attractor dynamics (no noise). The blue line shows the error without a discrepancy model, and the black dashed line shows the error with a discrepancy model recovering the missing physics.}
    \label{fig:Lorenz_IDphysics}
    \vspace{2cm}
\end{figure}

\begin{figure}[t]
    \centering
    \includegraphics[width=0.95\textwidth]{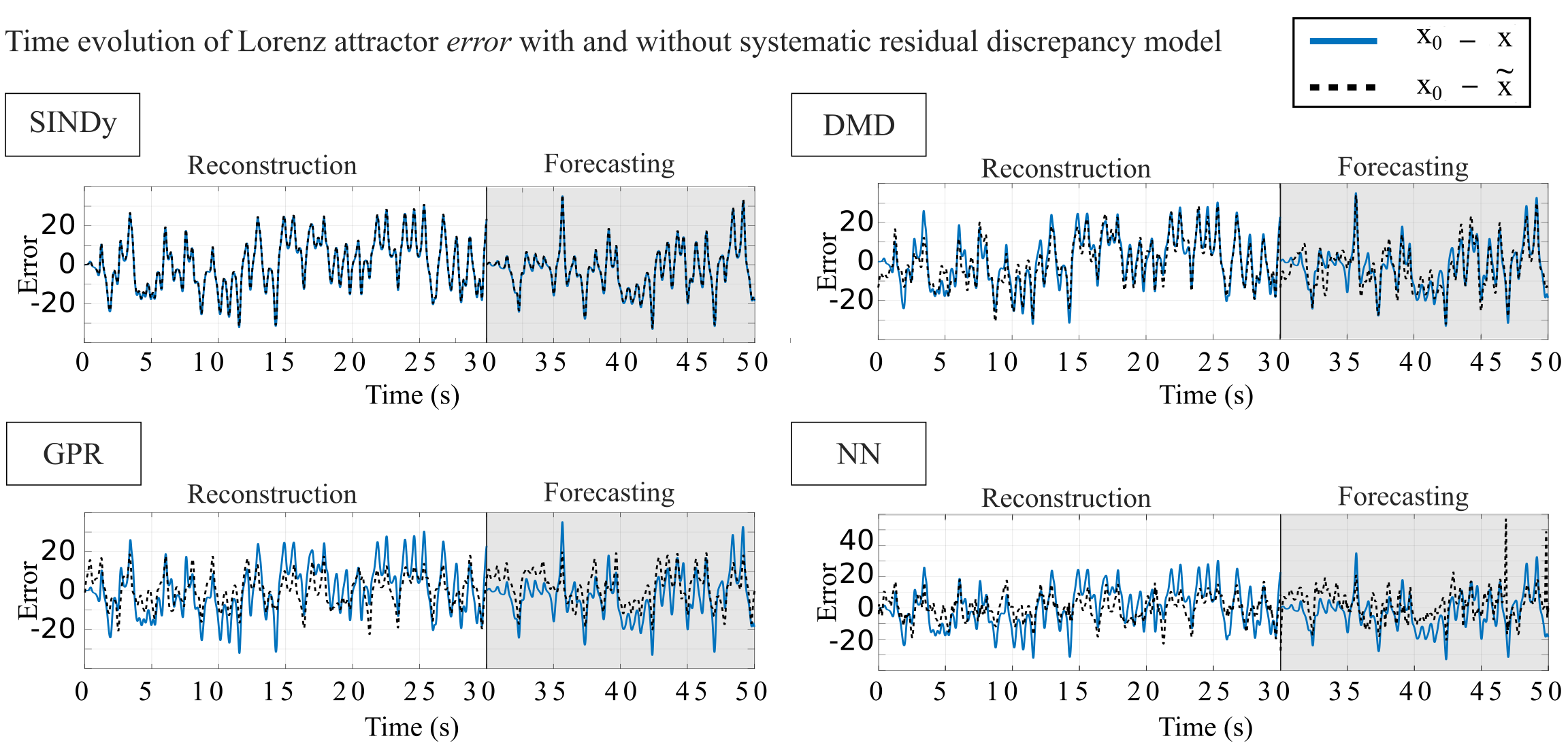}
    \caption{Remaining state-space error with and without a discrepancy model of the state-space systematic residual to correct the approximate Lorenz attractor state-space solution (no noise). The blue line shows the error without a discrepancy model, and the black dashed line shows the error with a discrepancy model of the systematic residual. All model discovery methods struggled to learn the time evolution of this residual.}
    \label{fig:Lorenz_errorModel}
\end{figure}

\begin{figure}[h]
    \centering
    \includegraphics[width=\textwidth]{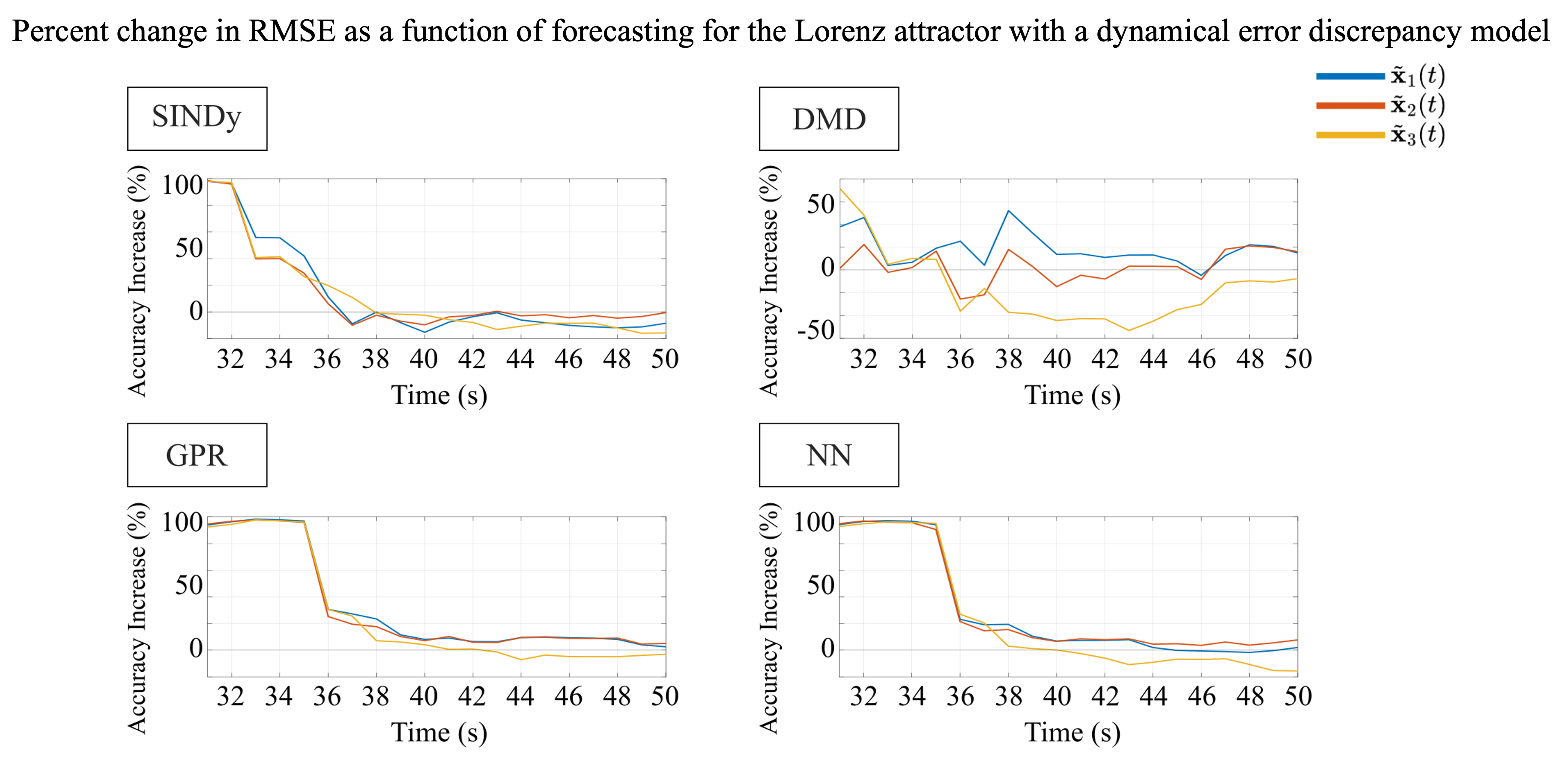}
    \caption{Percent change in RMSE as a function of forecasting. The discrepancy model for learning deterministic dynamical error in the Lorenz attractor example (no noise) demonstrates the relationship between the accuracy increase and the forecasting window; here, we see a decrease in accuracy with discrepancy model augmentation as the forecasting window is extended and settle near zero percent change from the approximate model. Note: the drop in forecasting RMSE for NN, GPR and arguably the SINDy discrepancy models corresponds with the first attractor 'jump' of the Lorenz. Predicting Lorenz bifurcations continues to be a challenging task in the field of dynamical systems.}
    \label{fig:TimeWindow}
\end{figure}

\subsubsection{Deterministic Dynamical Error}
We evaluated the ability of discrepancy modeling to recover missing physics for the Lorenz attractor. As seen in Fig.~\ref{fig:Lorenz_IDphysics}, the suite of model discovery methods had various success in learning the deterministic dynamical error for Lorenz. SINDy, GPR, and NN did well at learning the dynamical error, especially considering the sensitivity of chaotic systems to small dynamic deviations. These methods were able to reconstruct the first five seconds of the true Lorenz solution, as denoted by the zero-error in the reconstruction region (black dashed line). \textcolor{black}{Interestingly, the ability to reconstruct the true Lorenz behavior increased dramatically with increased measurement resolution ($\Delta t$).} In the forecasting region, SINDy had zero-error briefly but quickly became erroneous; this was most likely due to a small parameter deviation from the true dynamics that allowed the augmented dynamics to diverge from the true dynamics. Both GPR and NN were able to forecast the first five seconds of the true dynamics. In both the reconstruction and forecasting regions, the jump in non-zero error (blue line) corresponds with the attractor jump of the Lorenz dynamics. Unsurprisingly, DMD --- a linear model discovery method --- fails to learn the deterministic error for the Lorenz attractor example. Innovations from DMD's base code package have addressed some of the dynamical challenges due to nonlinearity and chaos, and continued innovation may allow for recovery of missing physics~\cite{brunton2017chaos}.

\begin{figure}[t]
    \centering
    \includegraphics[width=\textwidth]{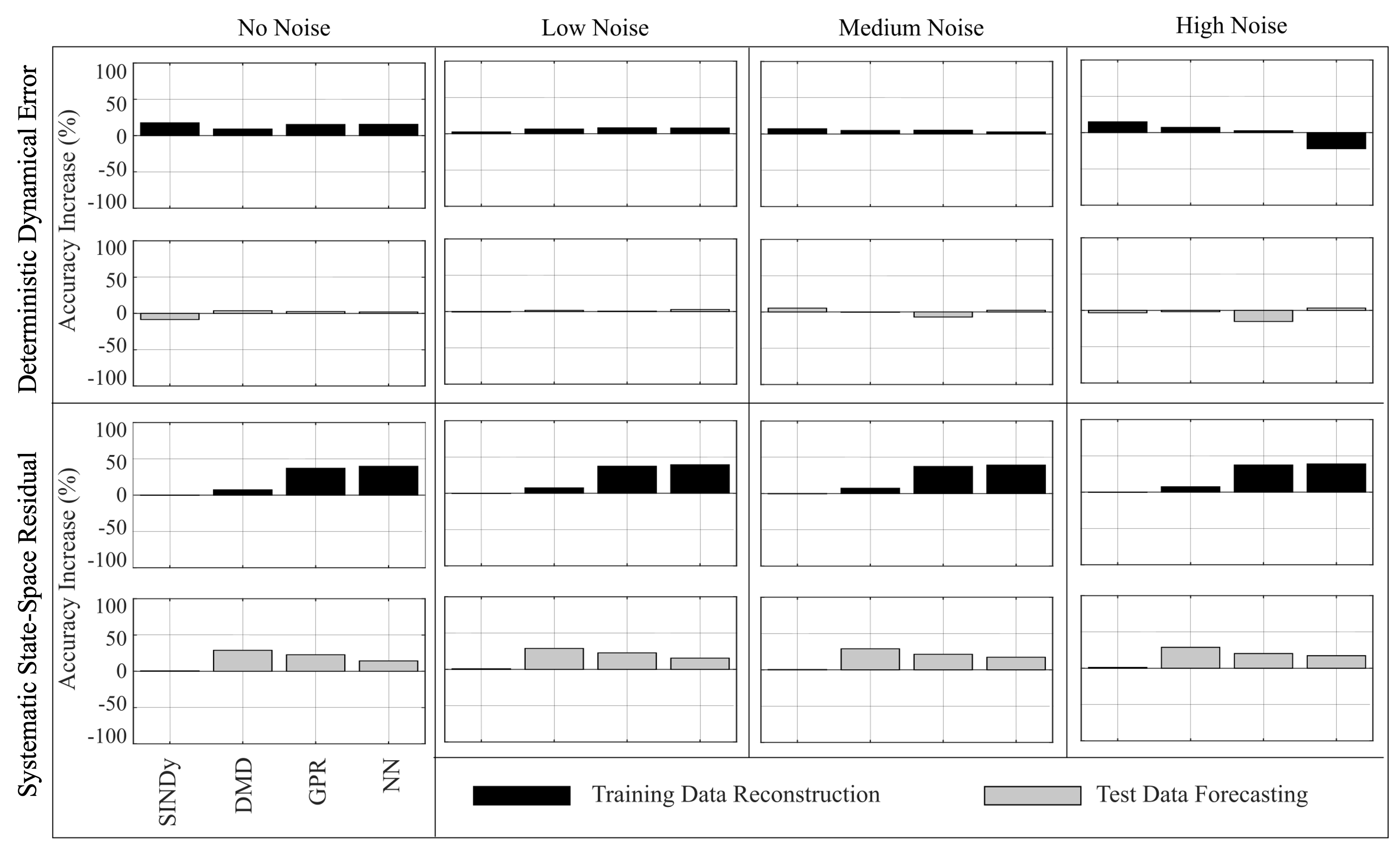}
    \caption{Increase in accuracy with discrepancy modeling augmentation in Lorenz attractor. Increase in \% accuracy is calculated as the root mean squared error (RMSE) between true and augmented state space solutions for reconstruction of the training region and forecasting in the test region. Results are shown for no (0\%), low (0.1\%), medium (1\%), and high (10\%) levels of noise. Discrepancy modeling shows promise for overcoming model-measurement mismatch, even for nonlinear and chaotic systems such as the Lorenz attractor.}
    \label{fig:Lorenz_RMSE_Compare}
\end{figure}

\subsubsection{Systematic State-Space Error}
We next evaluated the ability of discrepancy modeling to learn the evolution of the systematic state-space residual for the Lorenz attractor. As seen in Fig.~\ref{fig:Lorenz_errorModel}, the suite of model discovery methods struggled to model the time evolution of the residual. Similar to Van der Pol results above, SINDy was unable to learn a discrepancy model the state-space residual for the Lorenz attractor; the augmented state-space solution showed no change in time series solution as compared to the approximate state-space solution. Similarly, DMD showed minimal impact on the approximate state-space solution when corrected using the learned state-space residual discrepancy model. GPR and NN exhibited reduced error in the augmented state-space solutions as compared to the approximate; however, it does not appear that a discrepancy model of the residual alone can recover the true Lorenz attractor solution. Using discrepancy modeling to learn the state-space residual may benefit from improved data quantity, as well as data assimilation techniques to combat known challenges with chaotic deterministic systems~\cite{evensen1997advanced, gauthier1992chaos, miller1994advanced}. 

\subsubsection{Adding Noise}
We re-ran both discrepancy modeling approaches with the Lorenz attractor and increasing levels of Gaussian noise added to the 'true' Lorenz dynamics to evaluate the effect of sensor measurement \textit{quality}. We use noise level of $\sigma = [0.1\%, \; 1\%, \; 10\%] $. As seen in Fig.~\ref{fig:Lorenz_RMSE_Compare}, discrepancy modeling shows promise for overcoming model-measurement mismatch, even for highly nonlinear, chaotic systems such as the Lorenz attractor. Firstly, despite seeing an increase in accuracy as compared to the approximate model in learning a discrepancy model of the systematic state-space residual, the model discovery method were unable to reliably predict the attractor 'jump', which is a known challenge for data-driven modeling of chaotic dynamical systems~\cite{brunton2017chaos}. Secondly, we noticed a relationship between the accuracy increase and the forecasting window, as seen in Fig.~\ref{fig:TimeWindow}. While discrepancy modeling appears to have little impact on reducing the model-measurement mismatch, the percent change in error in Fig.~\ref{fig:Lorenz_RMSE_Compare} was computed over the entire forecasting window. However, we see that percent change in root mean squared error (RMSE) is a function of forecasting. Accuracy increase with discrepancy model augmentation begins near 100\% and decreases as the forecasting window is extended, eventually settling near zero percent change from the approximate model. Note the drop in RMSE around five seconds of forecasting for NN, GPR, and arguably the SINDy discrepancy models corresponds with the first attractor 'jump' of the Lorenz.

\subsection{Burgers' Equation: Spatio-temporal Nonlinearity}
Finally, we evaluate discrepancy modeling with a canonical system in partial differential equations. We applied the findings from our previous examples from the Van der Pol oscillator and the Lorenz attractor to inform our implementation of discrepancy modeling for the Burgers' equation. In this 'scenario', our intent for discrepancy modeling is for rapid evaluation with interpretability. Additionally, we have good data quality (no noise) and low data quantity (one set of spatio-temporal snapshots). Therefore, we chose to use DMD to learn a discrepancy model of the missing physics. We simulated Burgers' equation, which contains increasingly complex dynamical features, such as spatio-temporal nonlinearity:
\begin{equation}
    \frac{\partial u}{\partial t} + u\frac{\partial u}{\partial x} = \nu \frac{\partial^2 u}{\partial x^2}
    \label{eq:Burgers_platonic}
\end{equation}
using $\nu = 0.1$ $t = [0, 50]$ and $\Delta t = 0.01$, to generate our Platonic or approximate dynamics. To this system, we again added an $\epsilon$-small nonlinearity:
\begin{equation}
    \frac{\partial u_{0}}{\partial t} + u_{0}\frac{\partial u_{0}}{\partial x} + \epsilon u_{0}^3 = \nu \frac{\partial^2 u_{0}}{\partial x^2}
    \label{eq:Burgers_truth}
\end{equation}
and simulate to generate the 'true' system behavior using $\epsilon = 0.01$. To create the data matrix $x \in \mathcal{R}^{1xn}$, Eqs.~\ref{eq:Burgers_platonic} and \ref{eq:Burgers_truth} were evaluated on a gridspace comprising $n = 256$ equally spaced spatial points and initiated using \textcolor{black}{$u_{0}(0) = u(0) = e^{(x+2)^2}$}.

\begin{figure}[t]
    \centering
    \includegraphics[width=\textwidth]{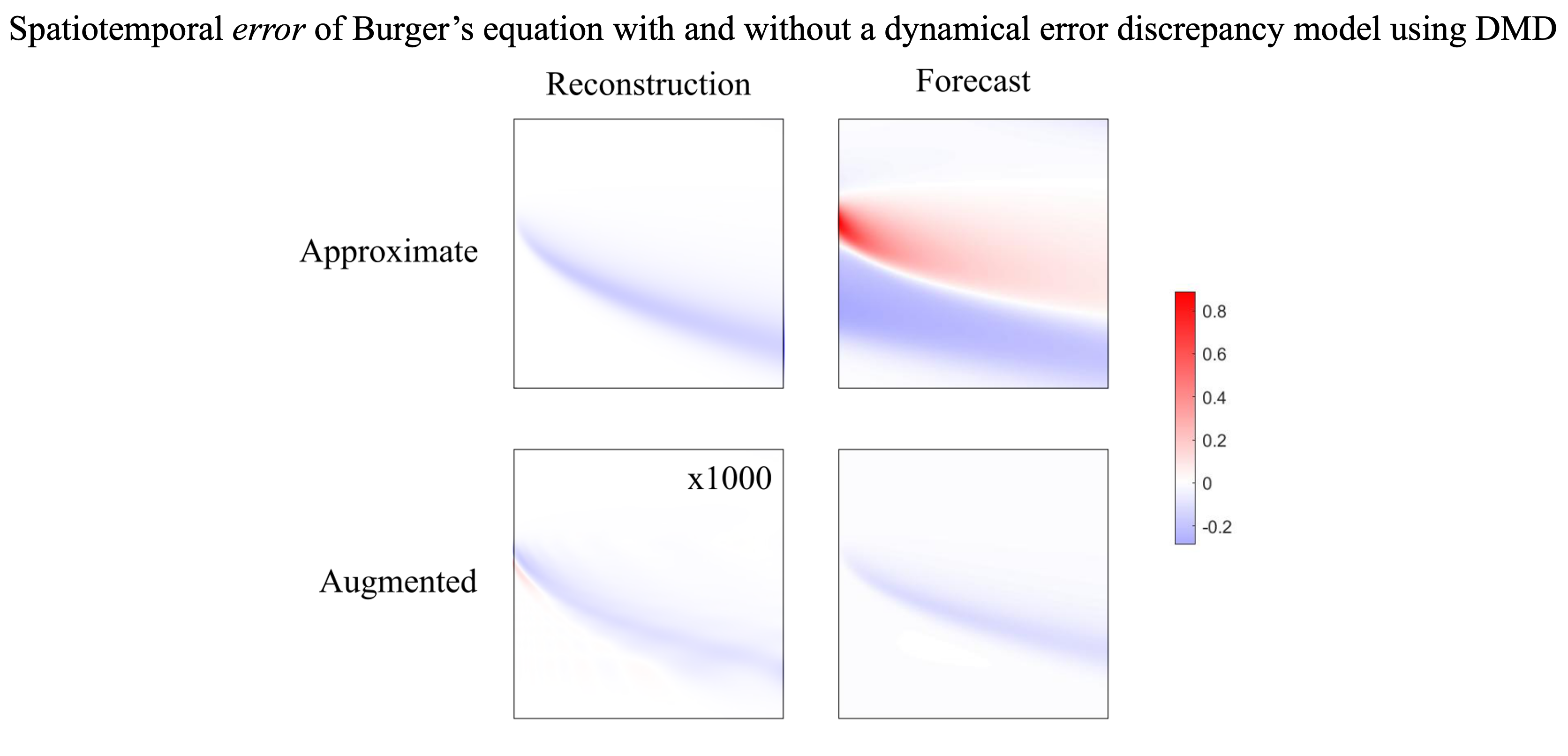}
    \caption{Remaining error with and without a discrepancy model of the deterministic dynamical error appended to the approximate Burgers' dynamics (no noise). No color (white) represents zero error as compared to the true system. Both red and blue denote non-zero error; the different colors only distinguish positive and negative error, respectively. Note that the augmented reconstruction error is multiplied by 1000. The dynamical error discrepancy model using DMD reconstructed true spatiotemporal dynamics with virtually zero remaining error and greatly diminished the error during forecasting. These results follow the same trends as seen in the Van der Pol example.}
    \label{fig:Burgers_errorModel}
\end{figure}

As seen in Fig.~\ref{fig:Burgers_errorModel}, DMD was successful in learning a discrepancy model recovering the missing physics in Burgers' equation. We note that trends from our non-chaotic ordinary differential equation example (Van der Pol) hold for this partial differential equation example (Burgers).

\subsection{Mass-Spring-Damper System}

One critique of this framework is that a systematic residual discrepancy model, as is formulated in Equation \ref{eq:correction}, is by construction a poor strategy for coping with imperfect dynamical models. \textcolor{black}{As is demonstrated in the three previous examples, an imperfect model is best improved by recovering the missing physics in the dynamical space instead of learning a discrepancy model in the state space. However, learning a discrepancy model of the state-space error is by construction a suitable approach for resolving an imperfect discrete dynamical system or for addressing state-space errors like unknown observations or measurement bias}. For example, a discrete dynamical system model for controls engineering could benefit from a discrepancy model updating the state space:

\begin{figure}[t!]
    \centering
    \includegraphics[width=\textwidth]{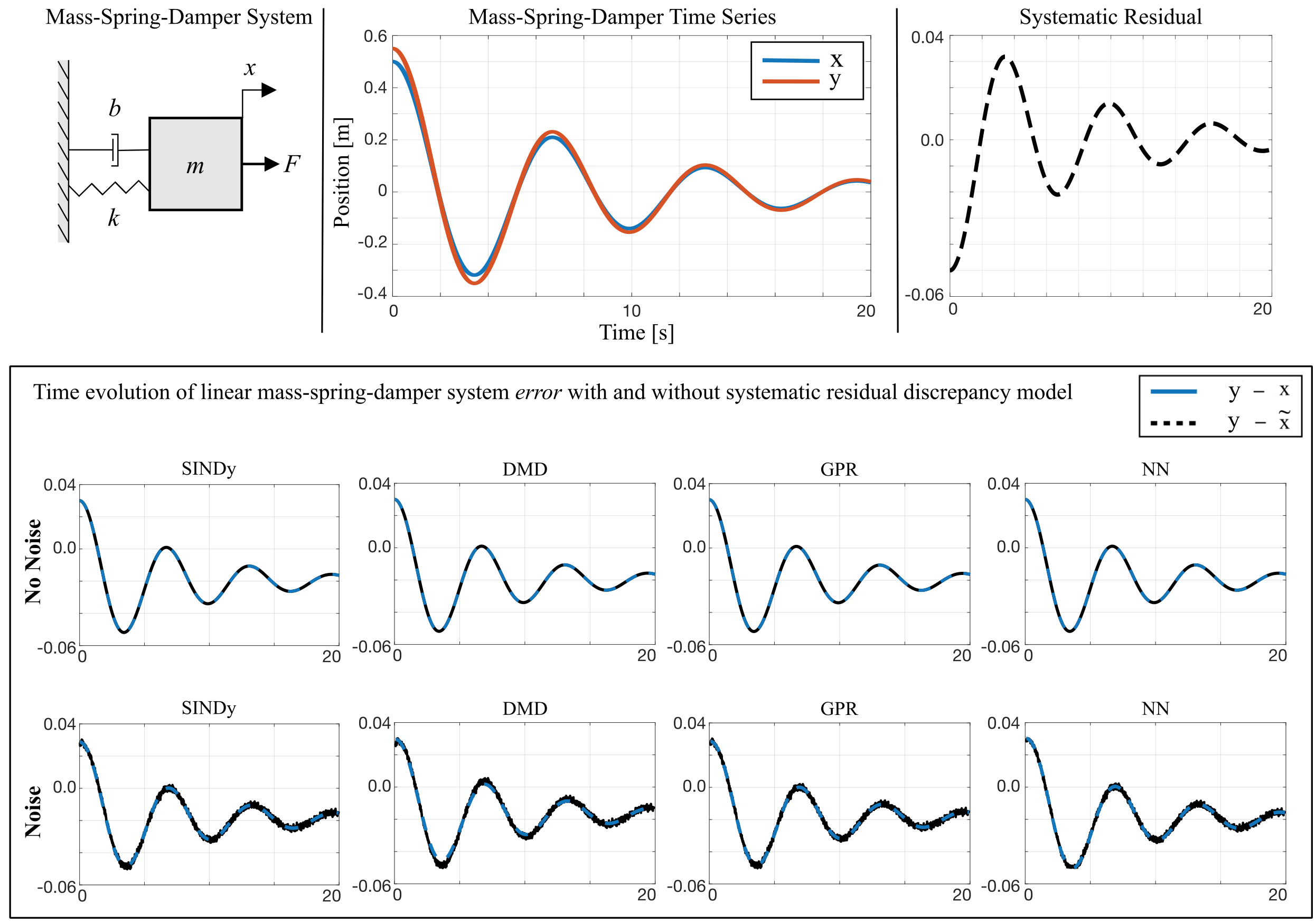}
    \caption{\textcolor{black}{(Top) A linear mass-spring-damper system with measurement bias, leading to a systematic state-space error. (Bottom) State-space error like measurement bias, as in this mass-spring-damper system, is well resolved using systematic residual discrepancy modeling, regardless of data-driven modeling method or noise. Importantly, this presents a neutral result: data assimilation like Kalman filtering may suffice to resolve \textcolor{black}{\textit{linear}} state-space observation error when a model of the true dynamics are known.}}
    \label{fig:movingobject}
\end{figure}

\begin{equation}
    x_{k+1} = f(x_k) + \delta(x_k)
    \label{eq:controls_example}
\end{equation}

\textcolor{black}{
To further demonstrate where the systematic residual discrepancy modeling approach would succeed, we simulated the linear mass-spring-damper system:}
\begin{equation}
    m\frac{d^2x}{dt^2} + b\frac{dx}{dt} + kx = F(t),
    \label{eq:MSD_linear}
\end{equation}
\textcolor{black}{with an observation error in the form of measurement bias:} 
\begin{equation}
    y_k = Cx(t_k) + \mathcal{N}(\mu, \sigma),
    \label{eq:measurement_bias}
\end{equation}
\textcolor{black}{
where the measurement bias, $\epsilon = 0.01$, is applied linearly through the observation matrix $C = $[1+$\epsilon$; 0]. We use the parameters $m = 2$, $b = 0.5$, $k = 2$, $t = $[0, 20], $\Delta t = 0.01$, $x_0 = x(0) =$ [0.5, 0], and $F(t) = x^3$.}

\textcolor{black}{
As seen in Fig. \ref{fig:movingobject}, regardless which data-driven modeling method was used, the state-space measurement bias was accurately learned. Even with noisy and biased measurements, the state-space error was corrected. However, the numerical simulations shown for the linear mass-spring-damper -- while simplistic -- presents a neutral result for addressing systematic residuals: \textit{linear} state-space error could be addressed by a data assimilation framework, a standard technique that is well-founded and mature in its formulation. For example, the Kalman filter and its variants have long achieved state-of-the-art performance \cite{law_data-2015, kalman1960new} and perform well when a model of the underlying dynamics is readily available, as is assumed in our systematic residual scenario.
} \textcolor{black}{However, when the systematic residual is produced by nonlinear means, such as by unknown nonlinear observations, as in:
\begin{equation}
    y_k = C(x(t_k)) + \mathcal{N}(\mu, \sigma), 
    \label{eq:measurement_bias_NL}
\end{equation}
data-driven methods for discrepancy modeling to resolve model-measurement mismatch will provide superior performance over data assimilation methods (which often assume known and unbiased observation error
statistics) \cite{kennedy2001bayesian, kalnay2003atmospheric, majda2012filtering}.}

\section{Conclusions and Guidelines for Discrepancy Modeling (Framework)}
In conclusion, discrepancy modeling for learning missing physics, modeling systematic residuals, and disambiguating between deterministic and random effects emerged as an important framework for principled investigation of model-measurement mismatch in dynamical systems. 
{\color{black}
The methods advocated for provide a suite of algorithms that can be broadly used in almost any realistic system where data and models are jointly used, such as emerging digital twin technologies~\cite{tao2018digital,jones2020characterising}}.
By leveraging improved observations to automate the process of learning better models, we can improve the characterization of underlying system dynamics. The two discrepancy modeling approaches introduced are distinct, yet nuanced: \textcolor{black}{(i) learning missing physics in the dynamical space to improve the underlying model and (ii) estimating the systematic residual in the state space to correct the model’s state approximations.}
While an abundance of data-driven modeling methods exist, we adapted a suite of four model discovery methods to demonstrate the mathematical implications of discrepancy modeling for practical applications. \textcolor{black}{Each combination of approach and method for discrepancy modeling demonstrates the need for clearly defined goals and an understanding of constraints, such as those imposed by data collection or model accessibility.} Further, by taking a deeper look at the mathematical implementations of each data-driven modeling method, we can better anticipate the utility, suitability, and interpretability of learned discrepancy models. Indeed, certain model discovery methods are more appropriate than others in conjunction with the chosen discrepancy modeling approach based on the intended use of the learned model, as well as sensor capabilities (data quality, quantity, resolution).
\textcolor{black}{To summarize, the goal of discrepancy modeling is to resolve model-measurement mismatch in dynamical systems. If the true dynamics are unknown, \textit{i.e.}, an imperfect dynamical model, identifying missing physics is best treated by learning the discrepancy model in the dynamical space. However, circumstances may dictate learning a discrepancy model in the state space, such as with discrete dynamical systems. If the true dynamics are known yet model-measurement mismatch still exists, learning a discrepancy model of the systematic residual is best treated in the state space. Indeed, data-driven discrepancy modeling is important for state space errors produced by nonlinear means or when data assimilation assumptions (\textit{e.g.}, known and unbiased observation error statistics) are invalid.}

\textcolor{black}{Because this manuscript is one of the first to comprehensively investigate how to handle the discrepancy of missing physics (\textit{i.e.}, deterministic dynamical error) in practical applications, we made a couple assumptions to isolate only the effect of missing physics on discrepancy modeling. Firstly, we specifically studied a fully observed dynamical system. If only partial observations are available, additional techniques can be implemented. Autoencoders are a standard technique in machine/deep learning for learning a full state space from limited observations. For example, shallow neural networks have been used to reconstruct high-dimensional states from limited measurements \cite{erichson2020shallow, carter2021data, sahba2022wavefront}. Additionally, time delay embedding is a widely-used technique, originally established by Taken’s theorem, to characterize a latent dimension from incomplete measurements \cite{hirsh2021structured, bakarji2022discovering}. Second, in this manuscript we specifically address an erroneous model, not erroneous initial conditions. These are two important but distinct challenges in modeling dynamical systems. Even with the perfect initial condition, if the underlying model is incorrect (\textit{e.g.}, idealized or missing physics), the time series behavior will be wrong from observations. Concurrently, an imperfect initial condition with a perfect model will also alter the estimated system behavior. We specifically investigate discrepancy modeling of missing physics, to de-conflate effects of model error and initial conditions. It is important we isolate the effect of missing physics for discrepancy modeling because incorrect parameter estimation can affect reconstruction and forecasting accuracy of a dynamical system. Indeed, an incorrect parameter – even with the correct dynamical structure – can lead to inaccurate system behavior \cite{kaheman2019learning}.}

Demonstrated in this manuscript is discrepancy modeling for a number of canonical spatial and/or temporal systems. For each example system, we evaluate reconstruction and/or forecasting accuracy. In particular, we vary the signal-to-noise ratio to demonstrate the impact of random fluctuations on the ability to disambiguate between deterministic and random effects. An important implication of this work is that there is no 'silver bullet' to automatically \textcolor{black}{resolve model-measurement mismatch.} In fact, in certain cases when opposing priorities cannot be reconciled, not using discrepancy modeling may be most appropriate. Further, this work demonstrates the limitations of discrepancy modeling as a function of observation noise; deterministic effects may exist within the model-measurement mismatch, but if random effects dominate, learning a discrepancy model of missing physics will be impossible and require improved sensor technology.

We summarize the the effects of the various discrepancy modeling paradigms:
\newline
\newline
\noindent (i) {\bf SINDy:} While SINDy requires more training data and a high signal-to-noise ratio, it results in the recovery of parsimonious dynamics and thus is agnostic to nonlinearity strength in the discrepancy dynamics. SINDy has minimal computational costs and provides immense opportunity to improve engineering design through its interpretable approach to learning the governing dynamics. Additionally, SINDy provides the architecture to evaluate discrepancy model capability and parameter dependence of the learned discrepancy dynamics. 
\newline
\newline
\noindent (ii) {\bf Dynamic Mode Decomposition:}
DMD's strength is in its interpretable approach to rapid model evaluation. Further, DMD is a good discrepancy modeling method when the signal-to-noise is too low for SINDy and is not dominated by random fluctuations. As DMD is a linear data-driven modeling approach, a DMD discrepancy model may be sensitive to the nonlinearity strength of discrepancy dynamics. 
\newline
\newline
\noindent (iii) {\bf Gaussian process regression:}
GPR is a powerful non-parametric Bayesian approach to discrepancy modeling in no/low/medium noise regime and lower data quantity requirements. It performed as well as --- or, in certain cases, better than --- a NN, along with lower associated computational costs. \textcolor{black}{GPR is a universal approximator with a closed form solution. While SINDy and DMD provide convenient solutions, they are not universal function approximators; conversely, NNs are approximators, but do not have closed form solutions. \textcolor{black}{The challenges of slowness and scalability with GPR can be addressed by using low rank or Random Feature approximations of Gaussian processes \cite{rahimi2007random}}}. 
\newline
\newline
\noindent (iv) {\bf Neural networks:}
NNs --- a parametric modeling approach --- shine when data in capturing complicated functions and generalizing to non-local behavior. If one can afford it, data assimilation for correcting systematic state space residual, for example with a GPR or NN, will be highly effective. For example, in chaotic systems such as the Lorenz, data assimilation will provide a consistent state-space correction versus modeling systematic error without feedback \cite{brajard2021combining}.
\newline
\newline
\noindent {\bf Remark:} \textcolor{black}{With increased data resolution (\textit{e.g.}, a high capacity sensor), a discrepancy model learning the deterministic dynamical error can almost perfectly reconstruct and forecast over short-time both the Van der Pol and Lorenz dynamical systems highlighted in this manuscript (\textit{e.g.}, using a $\Delta t$ = 0.001 versus the 0.01 used for the results presented in this manuscript). This is particularly important for learning a dynamical discrepancy model using the general form of SINDy whose coefficient estimations of a sparse set of dynamical terms are sensitive to data resolution, quality, and quantity. Additional techniques such as statistical bagging and ensembling methods~\cite{fasel2021ensemble} can greatly improve SINDy's robustness with lower resolution; such statistical techniques have successfully improved the general form of DMD as well \cite{sashidhar2022bagging}. Indeed, each architecture (i)-(iv) used can be enhanced and improved~\cite{champion2019data, champion2020unified, rudy2017data,tu2013dynamic,rasmussen2003gaussian, narendra1992neural, brunton2016discovering, kutz2016dynamic, bagheri2014effects, askham2018variable, brunton2017chaos, de2020pysindy, swiler2020survey,kaheman2022automatic}; this comparative study does not claim to optimize the ability of each technique to perform at its absolute best.  Indeed, hyper-parameter tuning of each technique is problem specific and again depends upon the characteristics of measurement data and the intent of its use.  Rather, our goal is to demonstrate the various possibilities and their appropriate uses.  It is evident from the study that each method considered has strengths and weaknesses that are appropriate to consider depending upon data and intent.}
\newline
\newline
As data-driven modeling continues to gain momentum, it is imperative that researchers utilize domain knowledge (\textit{e.g.}, first principles physics) to model complex systems. However, in all disciplines, model-measurement mismatch exists. The lack of investigation into this error leaves the missed opportunity to resolve model-measurement mismatch, disambiguate deterministic effects, and improve the underlying model. 
\textcolor{black}{Importantly, there may exist a spectrum of other types of discrepancies, such as partial measurements, initial condition errors, stochastic processes, and non-Gaussian distributions — among others we may not be aware of yet. Discrepancy modeling for dynamical systems is in its infancy. While our comparative work is early and extensive, it is not exhaustive. Many open and exciting challenges exist for data-driven engineering of dynamical systems in practical engineering applications.}

\section*{Acknowledgements}  We are especially grateful to Kadierdan Kaheman and Steven Brunton for discussions related to discrepancy modeling.  MRE acknowledges support from the NSF under award GRFP DGE-1762114.  JNK acknowledges funding from the National Science Foundation AI Institute in Dynamic Systems grant number 2112085.

\section*{Code}
\textcolor{black}{
Code can be found at {{\em github.com/meganebers/Discrepancy-Modeling-Framework-code}}}.

%\newpage
\bibliographystyle{unsrt}
\bibliography{references}

\end{document}